\documentclass[11pt]{article}

\usepackage{ACL}
\usepackage{times}
\usepackage{latexsym}
\usepackage{graphicx}
\usepackage{algorithm}
\usepackage[T1]{fontenc}
\usepackage[utf8]{inputenc}
\usepackage{microtype}
\usepackage{algorithm}
\usepackage{algorithmic}
\usepackage{colortbl}
\usepackage{inconsolata}
\usepackage{multirow}  % 为了使用跨行表格加的
\usepackage{amssymb}   % 为了使用粗体R
\usepackage{graphicx}  % 使用图片所需要的包
\usepackage{amsmath}
\usepackage{color}
\usepackage{booktabs}

\usepackage{subfigure} % 为了使用子图
\usepackage{makecell} % 为了自定义表格长度
\usepackage{booktabs} % 为了使用三线表
\usepackage{enumitem} % 为了让项目符号取消缩进
\usepackage{multirow} % 为了让允许表格多行显示
\usepackage{booktabs} % 为了允许在合并后的内容内换行
\usepackage{stfloats}
\usepackage{cite}
\usepackage{inconsolata}

\usepackage{times}
\usepackage{latexsym}

\usepackage[T1]{fontenc}
\usepackage[utf8]{inputenc}

\usepackage{microtype}

\usepackage[utf8]{inputenc}
\usepackage{graphicx}
\usepackage{amsmath}
\usepackage{amsthm}
\usepackage{booktabs}
\usepackage{algorithm}
\usepackage{algorithmic}
\usepackage{amssymb}
\usepackage{multirow}
\usepackage{array}
\usepackage{setspace}
\usepackage{pifont}
\usepackage{adjustbox}
\usepackage{hyperref}
\usepackage{makecell}

\title{Improving the Robustness of Distantly-Supervised Named Entity \\ Recognition via Uncertainty-Aware Teacher Learning \\ and Student-Student Collaborative Learning}

\author{
Shuzheng Si$^{1, 2}$\footnotemark[1], 
Helan Hu$^{1, 2}$\footnotemark[1],
Haozhe Zhao$^{1, 2}$\footnotemark[1], \textbf{Shuang Zeng}$^{3}$ \\ 
 \textbf{Kaikai An}$^{1,2}$, \textbf{Zefan Cai}$^{1,2}$, and \textbf{Baobao Chang}$^{1, 4}$\footnotemark[2] 
\\
$^1$National Key Laboratory for Multimedia Information Processing, Peking University \\ 
$^2$School of Software and Microelectronics, Peking University 
$^3$Tencent Inc.\\
$^4$Jiangsu Collaborative Innovation Center for Language Ability, \\
Jiangsu Normal University, Xuzhou, China
}

\begin{document}
\maketitle

% !TeX program = xelatex 
\begin{abstract}
% Distantly-Supervised Named Entity Recognition (DS-NER) effectively alleviates the burden of annotation, 
% To alleviate the burden of annotation, 
Distantly-Supervised Named Entity Recognition (DS-NER) is widely used in real-world scenarios.
It can effectively alleviate the burden of annotation by matching entities in existing knowledge bases with snippets in the text but suffer from the label noise.
Recent works attempt to adopt the teacher-student framework to gradually refine the training labels and improve the overall robustness.
However, these teacher-student methods achieve limited performance because the poor calibration of the teacher network produces incorrectly pseudo-labeled samples, leading to error propagation.
Therefore, we propose:
(1) Uncertainty-Aware Teacher Learning that leverages the prediction uncertainty to reduce the number of incorrect pseudo labels in the self-training stage;
(2) Student-Student Collaborative Learning that allows the transfer of reliable labels between two student networks instead of indiscriminately relying on all pseudo labels from its teacher, and further enables a full exploration of mislabeled samples rather than simply filtering unreliable pseudo-labeled samples.
We evaluate our proposed method on five DS-NER datasets, demonstrating that our method is superior to the state-of-the-art DS-NER methods.
% which simultaneously allows the training set can be fully explored.

% allows the training set can be fully explored
\end{abstract}
\renewcommand{\thefootnote}{\fnsymbol{footnote}}
% \footnotetext[1]{Equal contribution, ordered alphabetically by the last name. Email: sishuzheng@stu.pku.edu.cn}
\footnotetext[1]{~Equal Contribution. Email: sishuzheng@stu.pku.edu.cn.}
\footnotetext[2]{~Corresponding Author.}
\renewcommand{\thefootnote}{\arabic{footnote}}

\section{Introduction}
\label{intro}
\noindent
% As a fundamental task in NLP, 
Named Entity Recognition (NER) aims to locate and classify named entities in text, which plays an important role in many applications such as dialogue systems \citep{li-zhao-2023-em, liu-etal-2023-one, si-etal-2022-mining, si2024spokenwoz}.
However, deep learning-based NER methods usually require a substantial quantity of high-quality annotation for training models, which is exceedingly costly.
% due to the token-level labeling.
% To alleviate the burden of annotation, 
Therefore, Distantly-Supervised Named Entity Recognition (DS-NER) is widely used in real-world scenarios, which can automatically generate massive labeled training data by matching entities in existing knowledge bases with snippets in text.

\begin{figure}
    \centering
    \includegraphics[width=7.5cm]{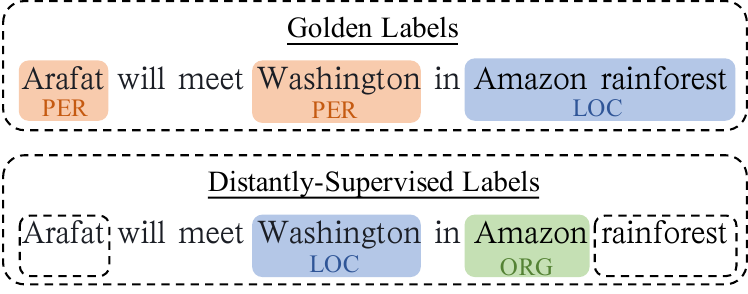}
    \caption{A sample generated by DS-NER. ``Amazon'' and ``Washington'' are inaccurate annotations. ``Arafat" and ``rainforest" are the incomplete annotations.}
    \label{fig_example}
\end{figure}

However, DS-NER suffers from two issues:
(1) \textbf{Incomplete Annotation}: due to the limited coverage of knowledge bases, many entity mentions in text cannot be matched and are wrongly labeled as non-entity,
and 
(2) \textbf{Inaccurate Annotation}:  the entity with multiple types in the knowledge bases may be labeled as an inaccurate type in the text, due to the context-free matching process.
As shown in Figure \ref{fig_example}, the entity types of "Washington" and "Amazon" are wrongly labeled owing to context-free matching, and "Arafat" is not recognized due to the limited coverage of resources.

Therefore, many works attempt to address these issues \citep{DBLP:conf/acl/PengXZFH19, zhou-etal-2022-distantly, DBLP:conf/iclr/LiL021, si-etal-2022-scl, si-etal-2023-santa}.
Recently, the self-training teacher-student framework in DS-NER has attracted increasing attention \citep{DBLP:conf/kdd/LiangYJEWZZ20, zhang-etal-2021-improving-distantly, DBLP:conf/aaai/QuZLWHZ23}, as it can handle inaccurate and incomplete labels simultaneously, and use generated pseudo labels to make full use of the mislabeled samples from DS-NER dataset.
% further reduce the negative effect of noisy labels.
This self-training framework firstly uses generated reliable pseudo labels from the teacher network to train the student network, and then updates a new teacher by shifting the weights of the trained student. 
Through this self-training loop, the training labels are gradually refined and model generalization can be improved.
Specifically, BOND \citep{DBLP:conf/kdd/LiangYJEWZZ20} designs a teacher-student network and selects high-confidence pseudo labels as reliable labels to get a more robust model.
SCDL \citep{zhang-etal-2021-improving} further improves the performance by jointly training two teacher-student networks, then selects consistent and high-confidence pseudo labels between two teachers as reliable labels.
ATSEN \citep{DBLP:conf/aaai/QuZLWHZ23} designs two teacher-student networks by considering both consistent and inconsistent high-confidence pseudo labels between two teachers and also proposes fine-grained teacher updating to achieve advanced performance.

The above teacher-student methods highly rely on using the high-confidence pseudo labels (e.g., pseudo labels with confidence values greater than 0.7) as reliable labels, as they assume that the teacher model's predictions with high confidence tend to be correct.
% and reliable.
However, this assumption may be far from reality.
Neural networks are usually poorly calibrated \citep{DBLP:conf/icml/GuoPSW17, DBLP:conf/iclr/RizveDRS21}, i.e., the probability associated with the predicted label usually reflects the bias of the teacher network and does not reflect the likelihood of its ground truth correctness.
Therefore, a poorly calibrated teacher network can easily generate incorrect pseudo labels with high confidence.
We argue that previous teacher-student methods achieve limited performance because poor network calibration produces incorrect pseudo-labeled samples, leading to error propagation.

We aim to reduce the effect of incorrect pseudo labels within the teacher-student framework by un\textbf{C}ertainty-aware t\textbf{E}acher a\textbf{N}d \textbf{S}tudent-Student c\textbf{O}llaborative lea\textbf{R}ning (\textbf{CENSOR}).
Specifically, we apply two teacher-student networks to provide multi-view predictions on training samples.
We propose Uncertainty-aware Teacher Learning that leverages the prediction uncertainty to guide the selection procedure of pseudo labels.
Then, we use both uncertainty and confidence as indicators to select pseudo labels, reducing the number of incorrect pseudo labels selected by confidence scores from poorly calibrated teacher networks.
We only select the pseudo labels with high confidence and low uncertainty as reliable labels, since these selected labels are more likely to contain less noise.
Subsequently, to further reduce the risk of learning incorrect pseudo labels and make a full exploration of mislabeled samples, we introduce Student-Student Collaborative Learning that allows the transfer of reliable labels between two student networks.
In each batch of data, each student network views its small-loss pseudo labels (e.g., pseudo labels of 10\% samples with the smallest loss) as reliable labels and then teaches such reliable labels to the other student network for updating the parameters.
In this way, a student network does not completely rely on all the pseudo labels from its poorly calibrated teacher network.
Meanwhile, different from just filtering unreliable pseudo-labeled samples, this component provides the opportunity for the incorrect pseudo-labeled samples to be correctly labeled by the other teacher-student network, allowing the full exploration of training data.
Experiments demonstrate that our method significantly outperforms previous methods, e.g., improving the F1 score by an average of 1.87\% on five DS-NER datasets.
% Further analysis shows that CENSOR reduces the percentage of incorrect pseudo labels in the training stage, and NITENDO is also robust to different noise ratios.
% Further analysis demonstrates the rationality and robustness of our proposed CENSOR.

\section{Task Definition}
\label{task_def}
\noindent
Given the training corpus ${D_{ds}}$ where each sample ($x_i$, $y_i$), $x_i$ represents {\it i}-th token, and $y_i$ is the label. 
Each entity is a span of the text, associated with an entity type.
We use the BIO scheme for sequence labeling.
% as \citet{DBLP:conf/aaai/QuZLWHZ23}. 
The beginning token of an entity is labeled as \textit{B-type}, and others are \textit{I-type}. 
The non-entity tokens are labeled as \textit{O}.
Traditional NER is a supervised learning task based on a clean dataset. 
We focus on the practical scenario where the training labels are noisy due to distant supervision, i.e., the revealed tag $y_i$ may not correspond to the underlying correct one. 
Thus, the challenge of DS-NER is to reduce the negative effect of noisy annotations.

\begin{figure*}
    \centering
    % \vspace{-7mm}
    \includegraphics[scale=0.5]{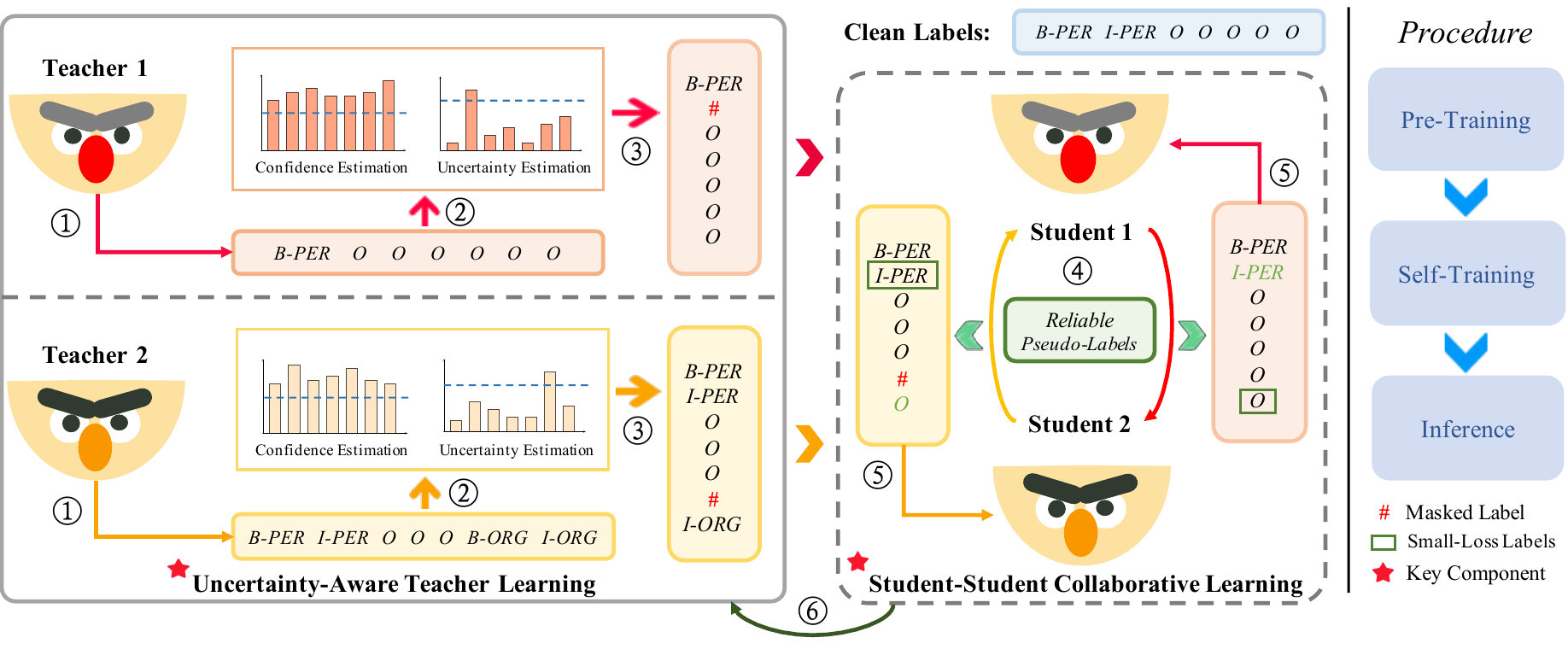}
    \caption{General architecture of CENSOR, which consists of two teacher-student networks. 
    [\ding{172}] means the teacher network first generates pseudo labels. 
    [\ding{173}] means estimating the confidence and uncertainty of generated pseudo labels.
    [\ding{174}] means selecting reliable pseudo labels according to confidence and uncertainty, where masked pseudo labels will not be used to update the student network.
    [\ding{175}] means using Student-Student Collaborative Learning to transfer the reliable pseudo labels.
    [\ding{176}] means using selected reliable pseudo labels to update the corresponding student network.
    [\ding{177}] means updating a new teacher by shifting the weights of the trained student.
    }
    \label{fig_model}
\end{figure*}

\section{Methodology}
As shown in Figure \ref{fig_model}, CENSOR consists of two teacher-student networks to handle the noisy label. 
To avoid overfitting the incorrect pseudo labels generated by poorly calibrated teacher networks, we introduce Uncertainty-Aware Teacher Learning that leverages the prediction uncertainty to guide the label selection.
We also propose Student-Student Collaborative Learning that allows reliable label transfer between two student networks, further reducing the risk of learning incorrect pseudo labels and making a full use of mislabeled samples.

\subsection{Teacher-student Framework}
Neural networks excel at memorization \citep{DBLP:conf/icml/ArpitJBKBKMFCBL17}.
However, when noisy labels become prominent, deep-learning-based NER models inevitably overfit noisy labeled data, resulting in poor performance. 
The purpose of the teacher-student methods is to select reliable labels (i.e., pseudo labels that are more likely to be labeled correctly), to reduce the negative effect of label noise. 
Self-training involves the teacher-student network, where the teacher network first generates pseudo labels to participate in label selection.
Then the student is optimized via back-propagation based on selected reliable labels, and the teacher is updated by gradually shifting the weights of the student with an exponential moving average (EMA).
Following \citet{DBLP:conf/aaai/QuZLWHZ23}, we train two sets of teacher-student networks using two different NER models to provide multi-view predictions on training samples.

\subsection{Uncertainty-Aware Teacher Learning}
In the DS-NER task, one of the main challenges of the teacher-student framework is to evaluate the correctness of the generated pseudo labels of the teacher model.
Previous methods \citep{DBLP:conf/kdd/LiangYJEWZZ20, zhang-etal-2021-improving-distantly,DBLP:conf/aaai/QuZLWHZ23} generally assume that high-confidence predictions tend to be correct.
Therefore, they select the samples with high-confidence pseudo labels (e.g., pseudo labels with confidence values greater than 0.7) as training data. 
However, the teacher network is prone to generating high-confidence yet incorrect pseudo labels due to the poor calibration \citep{DBLP:conf/icml/GuoPSW17}.
This overconfidence is indicative of model bias rather than the true likelihood of correctness.
Therefore, relying solely on the teacher network's confidence as the indicator may not efficiently evaluate the correctness of the pseudo labels.

Meanwhile, we observe that when the NER model performs supervised learning on a mislabeled token, it receives two types of supervision from the incorrect label of the mislabeled token and the labels of semantically similar but correctly labeled tokens.
For example, ``Washington'' in Figure \ref{intro} is mislabeled as ``LOC'' (location), and the model trained with it tends to predict ``Washington'' as ``LOC'' instead of ``PER'' (person).
The model is also exposed to semantically similar but correctly labeled tokens, such as the token ``James'' labeled as ``PER'' in the training sentence ``U.S. President will meet James at the White House'', thus the model may also learn to generalize "Washington" as a ``PER''.
The knowledge in both types of supervision is eventually learned and saved to the network neurons.
However, as the training continues, the deep-learning-based model inevitably overfits the noisy labels due to its memorization capability \citep{DBLP:conf/icml/ArpitJBKBKMFCBL17}, rather than utilizing the correct knowledge learned from the labels of semantically similar but correctly labeled tokens.

\paragraph{Uncertainty Estimation}
Based on our observation, we find that randomly deactivating neurons introduces variability in predicted confidence of the incorrect pseudo label, which can be attributed to varying subsets of active neurons influencing each prediction.
Specifically, the randomness of deactivation of the network neurons makes the remaining network neurons sometimes retain more knowledge learned from the incorrect label of the mislabeled token, and sometimes retain more knowledge learned from the labels of semantically similar but correctly labeled tokens.
Consequently, such discrepancies can lead to inconsistencies in multiple predictions.
For the correctly labeled tokens, since their labels are the same as those of semantically similar tokens, the two types of knowledge stored in the network neurons are more consistent, so the predictions from the different subsets of active neurons tend to be more consistent.
Thus, we define the inconsistency of predictions from sampled teacher network neurons as uncertainty and evaluate the correctness of the generated pseudo labels.

Specifically, given the new input token $x^*$ and the pseudo label $\hat{y}^*$ generated by the teacher network $W$, we perform $K$ forward passes with Dropouts \citep{DBLP:conf/nips/KrizhevskySH12} through our teacher networks at inference time. 
In each pass, pre-defined parts of network neurons are randomly deactivated.
Then, we could yield $K$ subsets of active neurons $\{\hat{W}_1, \hat{W}_2, ..., \hat{W}_{K} \}$.
To estimate the uncertainty for each token in the sequence labeling task, we leverage the variance of the model outputs for each token from multiple forward passes:

\begin{equation}
\small
\begin{aligned}
s_{un}(y^* = \hat{y}^*|W, x^*) = Var[p(y^* = \hat{y}^*|\hat{W_k}, x^*)]^{K}_{k=1},
\end{aligned}
\end{equation}
where $Var[.]$ is the variance of distribution over the $K$ passes through the teacher network.
% Accordingly, the teacher network is aware of its uncertainty over each sample and how likely the label is noisy.
The lower uncertainty indicates the predictions from sampled teacher network neurons and the learned knowledge are more consistent, thus the pseudo label is more likely to be correct.

% teacher network is certain that the current pseudo-labeled sample has been well learned and the pseudo label is more likely to be correct.

\paragraph{Uncertainty-Aware Label Selection}
% Without any prior knowledge about which tokens are mislabeled, it is challenging to automatically detect them. 
Different from previous teacher-student methods only using confidence as the indicator to select reliable pseudo labels, we jointly consider the confidence and uncertainty in label selection.
For the confidence of the pseudo label $\hat{y}^*$, as follows:
\begin{equation}
\small
\begin{aligned}
\hat{y}^* &= argmax(p(y^*|W,x^*)) \\
s_{co}(y^* &= \hat{y}^*|W, x^*) = p(y^* = \hat{y}^*|W, x^*)
\end{aligned}
\end{equation}
A higher confidence value $s_{co}$ means the model is more confident for the pseudo label $\hat{y}^*$.
However, many of these selected pseudo labels with high confidence are also incorrect due to the poorly calibrated teacher network \citep{DBLP:conf/icml/GuoPSW17}, leading to error propagation in the self-training.
To reduce the effect of incorrect pseudo labels, we additionally use uncertainty score $s_{un}$ as the indicator.
Specifically, we select a subset of pseudo labels which are both high-confidence and low-uncertainty as reliable labels, since jointly considering confidence and uncertainty can further filter the incorrect pseudo labels with high confidence.
Thus, we define a masked matrix, i.e.,
\begin{equation}
\small
\begin{aligned}
M_{x^*} = \left\{
        \begin{array}{rcl}
        1     & s_{un} < \sigma_{ua}\quad and\quad s_{co} >\sigma_{co}; \\
        \\
        0     & Otherwise; \\
        \end{array} 
    \right.
\label{eql:mask_matrix}
\end{aligned}
\end{equation}
When $M = 0$, it means the pseudo-label may be incorrect and the sample should be masked in the self-training. $\sigma_{co}$ and $\sigma_{ua}$ are hyperparameters.

\subsection{Student-Student Collaborative Learning}
Based on Uncertainty-Aware Teacher Learning, the teacher network can utilize the correctly pseudo-labeled samples to alleviate the negative effect of label noise.
However, simply masking unreliable pseudo-labeled samples can lead to underutilization of the training set, as there is no chance for the incorrect pseudo-labeled samples to be corrected and further learned. 
Intuitively, if we can correct the incorrect pseudo label with the correct one, it will become a useful training sample.
Therefore, to address these shortcomings and incorporate Uncertainty-Aware Teacher Learning to make the teacher-student network more effective, we propose Student-Student Collaborative Learning.

The idea of Student-Student Collaborative Learning is to utilize two different student networks and let them learn from each other.
We regard small-loss samples as clean samples for training, in each batch of data, each student network views its small-loss pseudo labels (e.g., pseudo labels of 10\% samples with the smallest loss) as the reliable labels, and transfers such reliable labels to another student network for updating the parameters. 
% 1
These small-loss samples are far from the decision boundaries of the two models and thus are more likely to be true positives and true negatives \citep{feng-etal-2019-learning}.
% 1
In this way, a student network is able to not completely rely on all pseudo labels from the teacher network, further reducing the risk of learning incorrect pseudo labels generated by the poorly calibrated teacher network.
Moreover, the two different student networks may have different decision boundaries and thus are good at recognizing different patterns in data. 
Different from simply masking unreliable pseudo-labeled samples, this component also provides the opportunity for the incorrect pseudo-labeled samples to be correctly labeled by the other teacher-student network to make full use of the training data.

% one student model may guide the other on how to pick correctly pseudo-labeled samples and circumvent noise. 

% Second, the two student networks with different architectures may have different decision boundaries and thus are good at recognizing different patterns in data. 
% This may allow student network to help the other rectify errors in learning.

Specifically, for two student networks $s_1, s_2 $ and their parameters $W_{s_1}, W_{s_2}$, we first let $s_1$ (resp. $s_2$) select a small ratio of samples in this batch of data $\hat{D}$ that have small training loss.
For these selected samples $\hat{D}_{s_1}$ (resp. $\hat{D}_{s_2}$) from $s_1$ (resp. $s_2$), we use the corresponding generated pseudo labels $\hat{Y}_{s_1}$ (resp. $\hat{Y}_{s_2}$) as reliable labels and transfer such reliable labels to the other student network $s_2$ (resp. $s_1$) for updating the parameters $W_2$ (resp. $W_1$). 
The ratio of transferred labels is controlled by hyperparameter $\delta$.
% , e.g., $\delta=0.2$ means transferring 20\% small-loss labels.
In this way, two student networks can learn from each other's reliable labels, reducing the risk of learning from incorrect pseudo labels and making full use of the training data.

\subsection{Training and Inference}
% In this subsection, we introduce the procedure of CENSOR. 
Algorithm \ref{tab_code} in Appendix \ref{appendix_pseudocode} gives the pseudocode. 
The process can be divided into three stages: the pre-training, the self-training, and the inference.
\paragraph{Pre-Training Stage} We warm up two different NER models $W_{A}$ and $W_{B}$ on the noisy DS-NER dataset to obtain a better initialization, and then duplicate the parameters $W$ for both the teacher $W_{t}$ and the student $W_{s}$ (i.e., $W_{t_1}$= ${W_{s_1}}$= $W_{A}$, $W_{t_2}$= $W_{s_2}$= $W_B$). The training objective function is the cross entropy loss with the following form:
\begin{align}
    % \mathcal{L}(\theta)=-\frac{1}{MN}\sum_{i=1}^M \sum_{j=1}^N y_j^i {\rm{log}}(p(y_j^i|x_i;\theta))
    \mathcal{L}=-\frac{1}{N}\sum_{D_{ds}} y_i {\rm{log}}(p(y_i|W_s, x_i))    
\end{align}
where $y_i$ means the {\it i}-th token label of the {\it i}-th token $x_i$ in the DS-NER corpus $D_{ds}$ and $p(y_i|W_s, x_i)$ denotes its probability produced by student network $W_s$. {\it N} is the size of the training corpus.

\paragraph{Self-Training Stage}
In this stage, we select reliable pseudo-labeled tokens to train the two teacher-student networks respectively. 
Specifically, we select reliable labels generated by teachers $W_{t}$ and supervise the students $W_{s}$ with cross-entropy loss.
During the label selection, we use the proposed Uncertainty-Aware Label Selection to jointly consider the confidence and uncertainty as shown in Eq.~\ref{eql:mask_matrix} to reduce the effect of incorrect pseudo-labeled samples.
Meanwhile, we use Student-Student Collaborative Learning to allow student networks can learn from each other’s reliable labels by selecting the pseudo labels from small-loss samples.
Therefore, the training objective function of student networks $W_{s}$ in this stage is the cross entropy loss with the following form:
\begin{align}
    % \mathcal{L}(\theta)=-\frac{1}{MN}\sum_{i=1}^M \sum_{j=1}^N y_j^i {\rm{log}}(p(y_j^i|x_i;\theta))
    \mathcal{L}=-\frac{1}{N}\sum_{D_{ds}} M_i\hat{y}_i {\rm{log}}(p(\hat{y}_i|W_s, x_i))    
\end{align}
where $\hat{y}_i$ means the {\it i}-th pseudo-label generated by Student-Student Collaborative Learning and its teacher $W_{t}$.
$p(\hat{y}_i|W_s, x_i)$ denotes its probability produced by student network $W_s$ on generated pseudo-label. 
$M_i$ is indicator where the {\it i}-th token ${x}_i$ should be masked according to Eq.~\ref{eql:mask_matrix}.
Meanwhile, if $\hat{y}_i$ is the transferred pseudo-label from the other student, $M_i$ will be automatically set to 1 (unmasked).
That is, we are more inclined to trust judgments from the student model because the student network is updated earlier and more frequently than the teacher network, and therefore better able to capture the changes of pseudo labels.
{\it N} is the size of the training corpus.

Different from the optimization of the student network, we apply EMA as \citet{zhang-etal-2021-improving-distantly} to gradually update the parameters of the teacher:
% , as shown in Eq.~\ref{eql:ema}:
% , where $\alpha$ denotes the smoothing coefficient.
\begin{align}
    % \mathcal{L}(\theta)=-\frac{1}{MN}\sum_{i=1}^M \sum_{j=1}^N y_j^i {\rm{log}}(p(y_j^i|x_i;\theta))
    {W}_{t} \xleftarrow{} \alpha {W}_{t} + (1-\alpha) {W}_{s}
    \label{eql:ema}
\end{align}
where $\alpha$ denotes the smoothing coefficient.
With the conservative and ensemble properties, the usage of EMA has largely mitigated the bias. 
As a result, the teacher tends to generate more reliable pseudo labels, which can be used as new supervision signals in the denoising self-training stage.

\paragraph{Inference Stage}
In the inference stage, only the best model $W_{best}\in\{ W_{t_1}, W_{s_1}, W_{t_2}, W_{s_2}\}$ on the dev set is adopted for predicting the test data.
\begin{table*}[tb]
% \vspace{-5mm}
\renewcommand\arraystretch{1.2}
\centering
\resizebox{15.9cm}{3.6cm}{
\begin{tabular}{clccccccccccccccc}
\toprule
\multicolumn{2}{c}{\multirow{2}{*}{\textbf{Method}}} & \multicolumn{3}{c}{\textbf{CoNLL03}} & \multicolumn{3}{c}{\textbf{OntoNotes5.0}} & \multicolumn{3}{c}{\textbf{Webpage}} & \multicolumn{3}{c}{\textbf{Wikigold}} & \multicolumn{3}{c}{\textbf{Twitter}} \\ \cline{3-17} 
\multicolumn{2}{l}{} & \textbf{P} & \textbf{R} & \textbf{F1} & \textbf{P} & \textbf{R} & \textbf{F1} & \textbf{P} & \textbf{R} & \textbf{F1} & \textbf{P} & \textbf{R} & \textbf{F1} & \textbf{P} & \textbf{R} & \textbf{F1} \\ 
\midrule
% \multirow{2}{*}{(\romannumeral1)}
% & $\textbf{BiLSTM-CRF}$  & 91.35 & 91.06 & 91.21 & 85.99 & 86.36 & 86.17 & 50.07 & 54.76 & 52.34 & 55.40  & 54.30  & 54.90  & 60.01 & 46.16 & 52.18 \\
% & $\textbf{RoBERTa}$  & 89.14 & 91.10  & 90.11 & 84.59 &87.88 & 86.20  & 66.29 &79.73 & 72.39 &85.33 & 87.56 & 86.43 & 51.76 & 52.63 & 52.19 \\ \midrule
% \multirow{1}{*}{(\romannumeral1)}
& KB-Matching  & 81.13 & 63.75 & 71.40 & 63.86 & 55.71 & 59.51 & 62.59 &45.14 & 52.45 & 47.90 & 47.63 & 47.76 & 40.34 & 32.22 & 35.83 \\
\midrule
% \multirow{4}{*}{(\romannumeral2)}
% & \textbf{KB-Matching}  & 81.13 & 63.75 & 71.40 & 63.86 & 55.71 & 59.51 & 62.59 &45.14 & 52.45 & 47.90 & 47.63 & 47.76 & 40.34 & 32.22 & 35.83 \\
& BiLSTM-CRF & 75.50 & 49.10 & 59.50 & \textbf{68.44} & 64.50 & 66.41 & 58.05 & 34.59 & 43.34 & 47.55 & 39.11 & 42.92 & 46.91 & 14.18 & 21.77 \\
& DistilRoBERTa & 77.87 & 69.91 & 73.68 & 66.83 & 68.81 & 67.80 & 56.05  & 59.46  & 57.70  & 48.85  & 52.05 & 50.40 & 45.72 & 43.85 & 44.77 \\
& RoBERTa & 82.29       & 70.47       & 75.93       & 66.99       & 69.51       & 68.23       & 59.24       & 62.84       & 60.98       & 47.67       & 58.59       & 52.57       & 50.97       & 42.66       & 46.45      \\ \midrule
% \multirow{6}{*}{(\romannumeral3)}
& AutoNER & 75.21       & 60.40        & 67.00          & 64.63       & 69.95           & 67.18       & 48.82       & 54.23       & 51.39       & 43.54       & 52.35       & 47.54       & 43.26       & 18.69       & 26.10       \\
& LRNT   & 79.91       & 61.87       & 69.74       & 67.36       & 68.02       & 67.69       & 46.70        & 48.83       & 47.74       & 45.60        & 46.84       & 46.21       & 46.94       & 15.98       & 23.84      \\
& Co-teaching+          & 86.04       & 68.74       & 76.42       & 66.63       & 69.32       & 67.95       & 61.65       & 55.41       & 58.36       & 55.23       & 49.26       & 52.08       & 51.67       & 42.66       & 46.73      \\
& JoCoR  & 83.65       & 69.69       & 76.04       & 66.74       & 68.74       & 67.73       & 62.14       & 58.78       & 60.42       & 51.48       & 51.23       & 51.35       & 49.40        & {45.59}       & 47.42      \\
& NegSampling    & 80.17       & 77.72             & 78.93       & 64.59       & \textbf{72.39}       & 68.26        & {70.16}       & 58.78       & 63.97       & 49.49       & 55.35             & 52.26       & 50.25       & 44.95             & 47.45 \\
\midrule
& BOND    & 82.05       & {80.92}             & 81.48       & 67.14       & 69.61       & 68.35       & 67.37       & 64.19       & 65.74       & 53.44       & \textbf{68.58}             & 60.07       & 53.16       & 43.76             & 48.01       \\ 
% \midrule
& SCDL             & \textbf{87.96}             & 79.82       & 83.69             & \underline{67.49}       & 69.77       & \underline{68.61 }          & 68.71             & {68.24}             & {68.47}             & \underline{62.25}             & 66.12       & \underline{64.13}             & \textbf{59.87}             & 44.57       & {51.09}            \\ 
& ATSEN            & 85.75            & \underline{83.86}      & \underline{84.79}             & 65.69       & 70.71      & 68.11             & \underline{71.08}             & \underline{70.03}             & \underline{70.55}             & { 57.67}             &  54.71      & 56.15             & \underline{59.31}             & \underline{45.83}       & \underline{51.71}            \\
\midrule
% \rowcolor{blue!5}
\rowcolor{blue!5} & \textbf{CENSOR}             & \underline{87.33}            & \textbf{85.90}      & \textbf{86.61}             &  67.11      & \underline{71.01}     & \textbf{69.01}             & \textbf{75.89}            & \textbf{72.30}             & \textbf{74.05}             & \textbf{66.01}             &  \underline{68.10}      &  \textbf{67.05}             & 58.63            & \textbf{47.38}       & \textbf{52.41}    \\
\bottomrule 
\end{tabular}}
\caption{Main results on five DS-NER datasets. We report the baseline results from \citet{DBLP:conf/kdd/LiangYJEWZZ20, zhang-etal-2021-improving-distantly} and our experimental results with their official implementation in our devices.}
% (\romannumeral1) $\clubsuit$ marks the model trained on the fully clean dataset. (\romannumeral2) $\dag$ marks the model trained on noisy dataset without label denoising. (\romannumeral3) $\ddag$ marks the prior label denoising framework. $\star$ marks produced with official implementation.}
\label{tab_main}
\end{table*}

% \begin{table}
% \small
% \centering
% \renewcommand{\arraystretch}{1.1}
% \scriptsize
% \setlength{\tabcolsep}{2mm}{
% \begin{tabular}{lccc}
% % \hline
% \toprule
% \multirow{2}{*}{Method} & \multicolumn{3}{c}{Webpage}\\
% \cline{2-4} & \textbf{P}& \textbf{R}& \textbf{F1} \\
% \hline
% {Focal Loss}&{66.67}&{55.77}&{60.74} \\
% {CE}&{67.32}&{56.41}&{61.38} \\
% {GCE + SR}&{67.45}&{57.17}&{61.88} \\
% {CE \& GCE + SR}&{68.46}&{58.32}&{62.98}\\
% {\textbf{SANTA}}&{78.40}&{66.22}&{\textbf{71.79}}\\
% \hline
% % {w/o MFL}&{78.40}&{66.22}&{71.79}\\
% {w/o MFL }&{-}&{-}&{-}\\
% {\quad w. GCE + SR}&{77.78}&{61.49}&{68.68}\\
% {\quad w. CE}&{73.33}&{66.89}&{69.96}\\
% {w/o Memory Label Smoothing in MLF}&{74.02}&{63.51}&{68.36}\\
% {\quad w. Label Smoothing in MLF}&{74.05}&{65.54}&{69.53}\\
% \hline
% {w/o Entity-aware KNN}&{77.66}&{49.32}&{60.33}\\
% {\quad w. KNN-augmented Inference}&{67.94}&{60.14}&{63.80}\\
% {\quad w. Entity-aware CL}&{87.84}&{43.92}&{58.56}\\
% \hline
% {w/o GCE + SR} &{-}&{-}&{-}\\ 
% {\quad w. GCE}&{77.52}&{66.03}&{71.32}\\
% {\quad w. CE}&{76.56}&{66.22}&{71.01}\\
% {\quad w. MFL}&{72.31}&{66.43}&{69.25}\\
% \hline
% {w/o Boundary Mixup}&{70.71}&{47.30}&{56.68}\\
% {\quad w. Mixup}&{76.98}&{65.54}&{70.80}\\
% % \hline
% \toprule
% \end{tabular}
% }
% \caption{Ablation study on Webpage.}
% \label{tab_ablation} 
% \end{table}

\section{Experiment}
\label{sec:experiments}
% Compared with extensive baselines, CENSOR achieves significant improved performance in five DS-NER datasets.
% We also provide analyses to justify the effectiveness of CENSOR.

\subsection{Dataset}
\noindent
We conduct experiments on five DS-NER datasets, including CoNLL03 \citep{tjong-kim-sang-de-meulder-2003-introduction}, Webpage \citep{ratinov-roth-2009-design}, Wikigold \citep{DBLP:conf/acl-pwnlp/BalasuriyaRNMC09}, Twitter \citep{DBLP:conf/aclnut/GodinVNW15} and OntoNotes5.0 \citep{weischedel2013ontonotes}.
For the fair comparison, we follow the same knowledge bases and settings as \citet{DBLP:conf/kdd/LiangYJEWZZ20}, re-annotate the training set by distant supervision, and use the original dev and test set.
% and \citet{zhang-2022-improve}, and \citet{si-etal-2022-scl}.
% More datasets details can be found in Appendix \ref{appendix_dataset}.
% For CoNLL2003, OntoNotes5.0, Webpage, \citet{DBLP:conf/kdd/LiangYJEWZZ20} re-annotates the training set by distant supervision, and uses the original development and test set.
Statistics of datasets are shown in Appendix \ref{appendix_dataset}.

% For CoNLL2003, OntoNotes5.0, Webpage, \citet{DBLP:conf/kdd/LiangYJEWZZ20} reannotated the training set by distant supervision, and used the original development and test set.
% We keep the same knowledge bases as \citep{shang-etal-2018-learning} in BC5CDR.
% For EC, \citet{DBLP:conf/coling/YangCLHZ18} used the distant supervision to get training data, and labeled development and test set by crowd-sourcing.
% % Statistics of datasets are shown in Table \ref{tab_datasets}.

\subsection{Evaluation Metrics and Baselines}
\noindent
We use Precision (P), Recall (R), and F1 score as our evaluation metrics.
We compare CENSOR  with various baseline methods, including supervised methods and DS-NER methods. 
We also present the results of \textbf{KB-Matching}, which directly uses knowledge bases to annotate the test sets.

\paragraph{Supervised Methods}
We select \textbf{BiLSTM-CRF} \citep{ma-hovy-2016-end}, \textbf{RoBERTa} \citep{liu2019roberta} and \textbf{DistilRoBERTa} \citep{Sanh2019DistilBERTAD} as original supervised methods.
As trained on noisy DS-NER datasets, these methods achieve poor performance. 

\paragraph{DS-NER Methods}
We compare several DS-NER baselines.
% (1) methods that only consider the incomplete annotation:
% \textbf{Conf-MPU} \citep{zhou-etal-2022-distantly} uses multi-class PU-learning loss to better estimate the loss.
\textbf{AutoNER} \citep{shang-etal-2018-learning} modifies the standard CRF to get better performance under the noise. 
\textbf{LRNT} \citep{DBLP:conf/emnlp/CaoHCLJ19} leaves training data unexplored fully to reduce the negative effect of noisy labels.
\textbf{Co-teaching+} \citep{DBLP:conf/icml/Yu0YNTS19} and \textbf{JoCoR} \citep{DBLP:conf/cvpr/WeiFC020} are two classical collaborative learning methods to handle noisy labels in computer vision area.
% \textbf{NegSampling} \citep{DBLP:conf/iclr/LiL021} and \textbf{Weighted NegSampling} \citep{li-etal-2022-rethinking} uses down-sampling in non-entities to relief the misleading from incomplete annotation.
\textbf{NegSampling} \citep{DBLP:conf/iclr/LiL021} uses down-sampling in non-entities to relief the misleading from incomplete annotation.
% \textbf{SCL-RAI} \citep{si-etal-2022-scl} uses span-based supervised contrastive-learning loss and designed inference method to improve the robustness against incomplete annotation.
% (2) methods that consider two types of noise equally:
% \textbf{BOND} \citep{DBLP:conf/kdd/LiangYJEWZZ20} and \textbf{SCDL} \citep{zhang-etal-2021-improving} proposes teacher-student network to reduce the noise from distant labels.
% \textbf{ATSEN} 

\paragraph{Teacher-Student Methods for DS-NER}
Specifically, \textbf{BOND} \citep{DBLP:conf/kdd/LiangYJEWZZ20} designs a teacher-student network and selects high-confidence predictions as pseudo labels to get a robust model.
\textbf{SCDL} \citep{zhang-etal-2021-improving} improves the performance by training two teacher-student networks and selecting consistent high-confidence predictions between two teachers as pseudo labels.
\textbf{ATSEN} \citep{DBLP:conf/aaai/QuZLWHZ23} considers both consistent and inconsistent predictions with high confidence between two teachers and further proposes a fine-grained teacher updating method.
We report the results of ATSEN with official implementation in our devices.
% \textbf{RoSTER} \citep{meng-etal-2021-distantly} adopts GCE loss, self-training and noisy label removal step to improve the robustness.
% \textbf{CReDEL} \citep{ying-etal-2022-label} proposes an automatic distant label refinement model via contrastive-learning as a plug-in module.

\subsection{Experimental Settings}
Following \citet{DBLP:conf/aaai/QuZLWHZ23}, we adopt RoBERTa-base and DistilRoBERTa-base as two NER models for two teacher-student networks.
We use Adam \citep{DBLP:journals/corr/KingmaB14} as our optimizer.
We list detailed hyperparameters in the Appendix \ref{appendix_hyperparameter}.

\subsection{Main Results}
\label{sec:main_results}
\noindent
% Table \ref{tab_main} presents the main results of CENSOR.
Table \ref{tab_main} presents the performance of different methods measured by precision, recall, and F1 score.
Specifically, 
(1) CENSOR achieves new SOTA performance, showing superiority in the DS-NER task;
(2) Compared to original supervised methods, including BiLSTM-CRF, RoBERTa, and DistilRoBERTa, CENSOR improves the F1 score with an average increase of 23.04\%, 10.96\%, and 8.99\%, respectively, which demonstrates the necessity of DS-NER models and the effectiveness;
(3) Compared to classical de-noising methods in the computer vision area (e.g., Co-teaching+), simply using these methods can not achieve strong performance, since these methods were not initially designed for sequence labeling tasks and ignore the characteristics of the DS-NER task.
(4) Compared with teacher-student methods such as BOND, SCDL, and ATSEN, CENSOR achieves advanced performance, confirming that these teacher-student methods achieve limited performance because of the incorrect pseudo-labeled samples.
% calibration produces incorrectly pseudo-labeled samples and further introduce the error propagation.

\begin{table}[t]
  \centering
  \small
\resizebox{\linewidth}{!}{%
  \begin{tabular}{lccc}
    \toprule
    \textbf{Method}  & \textbf{P}  & \textbf{R}   & \textbf{F1}  \\ 
    \midrule
    \rowcolor{blue!5} \textbf{CENSOR} & \textbf{87.33}  & \textbf{85.90} & \textbf{86.61}  \\
    \midrule
\quad -w/o UTL & 86.56 (\textcolor{green!60!black}{-0.77})  & 84.37 (\textcolor{green!60!black}{-1.53}) & 85.45 (\textcolor{green!60!black}{-1.16})        \\
\quad -w/o SCL  & 86.44 (\textcolor{green!60!black}{-0.89})  & 83.98 (\textcolor{green!60!black}{-1.92}) & 85.19 (\textcolor{green!60!black}{-1.42})       \\
    \bottomrule 
  \end{tabular}
  }
  \caption{Ablation study on CoNLL03. UTL means Uncertainty-Aware Teacher Learning and SCL means Student-Student Collaborative Learning. }
  \label{tab:ablation}
  
\end{table}

\subsection{Analysis}

\paragraph{Ablation Study}
% To evaluate the effectiveness of each proposed component, we conduct the ablation study.
Shown in Table \ref{tab:ablation}, it is clear that Uncertainty-Aware Teacher Learning and Student-Student Collaborative Learning are both important to the model performance.
Removing each component can lead to a simultaneous decrease in precision and recall at the same time, showing that proposed components indeed improve performance.

\begin{figure}
    \centering
    \resizebox{\linewidth}{!}{%
        \includegraphics[width=7.5cm]{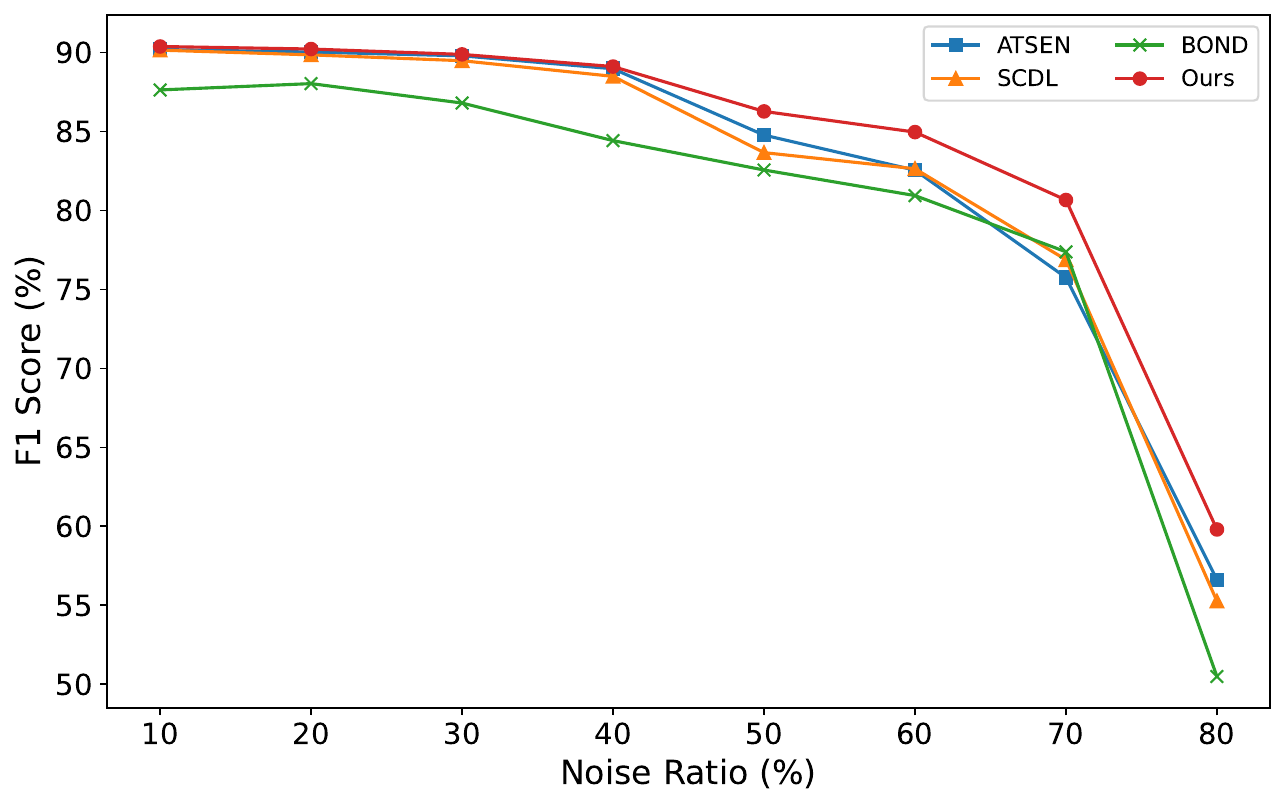}
    }
    \caption{F1 on CoNLL03 with different noise ratios.}
    \label{fig:noise}
\end{figure}

\paragraph{Robustness to Different Noise Ratios}
To investigate the robustness of the CENSOR in different noise ratios, we randomly replace $k$\% entity labels in the clean version (instead of the distantly-supervised version) of CoNLL03 training set with other entity types or non-entity.
In this way, we can construct different noise ratios of label noise and we further report the test F1 score on CoNLL03.
As shown in Figure \ref{fig:noise}, CENSOR achieves consistent advanced performance in different noise ratios, showing its satisfactory de-noising ability and strong robustness.
Meanwhile, when the noise ratio is above 50\%, CENSOR achieves more significant robustness, since CENSOR can select and generate more reliable labels due to the Uncertainty-Aware Teacher Learning and Student-Student Collaborative Learning from highly noisy data.
More detailed data can be found in Table \ref{tab_noise} in the Appendix.

\begin{table}[t]
  \centering
  \small
\resizebox{\linewidth}{!}{%
  \begin{tabular}{lccc}
    \toprule
    \textbf{Method}  & \textbf{P}  & \textbf{R}   & \textbf{F1}  \\ 
    \midrule
BOND & 80.87 (\textcolor{green!60!black}{-13.49})  & 78.04 (\textcolor{green!60!black}{- 7.09}) & 79.43 (\textcolor{green!60!black}{-10.08})        \\
SCDL  & 94.18 (\textcolor{green!60!black}{- 0.18})  & 77.11 (\textcolor{green!60!black}{- 8.02}) & 84.80 (\textcolor{green!60!black}{- 4.71})       \\
ATSEN  & {93.01} (\textcolor{green!60!black}{{- 1.35}})  & {82.96} (\textcolor{green!60!black}{- 2.17}) & {87.70} (\textcolor{green!60!black}{- 1.87})       \\
\midrule
\rowcolor{blue!5} \textbf{CENSOR} & \textbf{94.36}  & \textbf{85.13} & \textbf{89.51}  \\
\bottomrule 
  \end{tabular}
  }
\caption{Comparison of the effectiveness of reducing label noise on CoNLL03.}
  \label{tab:selection}
\end{table}

\paragraph{Effectiveness of Reducing Learned Noise}
To confirm previous teacher-student methods achieve limited performance because of incorrectly pseudo-labeled samples, we try to explore the effectiveness of reducing label noise from different teacher-student methods, including CENSOR, BOND, SCDL, ATSEN.
% Specifically, we report the best F1 score of selected reliable labels with ground truth labels in the clean version of CoNLL03 training set.
Specifically, we report the average F1 score of all selected (unmasked) pseudo labels for training during the self-training stage, using the labels from the clean version of the CoNLL03 training set as ground truth labels.
As shown in Table \ref{tab:selection}, CENSOR achieves a consistent advanced F1 score, which indicates CENSOR can select more correct labels based on Uncertainty-Aware Label Selection and Student-Student Collaborative Learning.
Thus, CENSOR can use more correct pseudo labels to update the parameters of student networks and further avoid error propagation, leading to outstanding overall performance on the test set.

\begin{table}[t]
  \centering
  \small
\resizebox{\linewidth}{!}{%
  \begin{tabular}{lccc}
    \toprule
    \textbf{Method}  & \textbf{P}  & \textbf{R}   & \textbf{F1}  \\ 
    \midrule
BOND & 80.42 (\textcolor{green!60!black}{-9.44})  & 76.46 (\textcolor{green!60!black}{-8.69}) & 78.39 (\textcolor{green!60!black}{-9.05})        \\
SCDL  & 87.42 (\textcolor{green!60!black}{-2.44})  & 75.85 (\textcolor{green!60!black}{-9.30}) & 81.22 (\textcolor{green!60!black}{-6.22})       \\
ATSEN  & {87.84} (\textcolor{green!60!black}{{-2.02}})  & {82.83} (\textcolor{green!60!black}{-2.32}) & {85.26} (\textcolor{green!60!black}{-2.18})       \\
\midrule
\rowcolor{blue!5} \textbf{CENSOR} & \textbf{89.86}  & \textbf{85.15} & \textbf{87.44}  \\
\bottomrule 
  \end{tabular}
  }
\caption{Comparison of teacher pseudo-labeling ability of different teacher-student methods on CoNLL03.}
  \label{tab:refine}
\end{table}

\paragraph{Effectiveness of Teacher Pseudo-labeling}
After confirming the effectiveness of reducing label noise, we attempt to further explore whether the teacher network could use more reliable labels to avoid error propagation, thus generating more correct pseudo labels.
As shown in Table \ref{tab:refine}, we report the best F1 score of teacher networks from different teacher-student methods on the clean version of CoNLL03 training set.
In detail, the teacher network from CENSOR correctly labels 87.44\% samples, achieving the most advanced precision, recall, and F1 score.
Compared to other teacher-student methods, including BOND, SCDL, and ATSEN, CENSOR improves the F1 score with an average increase of 9.05\%, 6.22\%, and 2.18\%, respectively, which demonstrates using more correct labels can avoid error propagation and make the teacher network generate more reliable labels.
In this way, the teacher network can make full use of the noisy samples in the DS-NER training set and help the teacher-student framework achieve outstanding performance on the test set.

\begin{figure}
    \centering
    \resizebox{\linewidth}{!}{%
        \includegraphics[width=7.5cm]{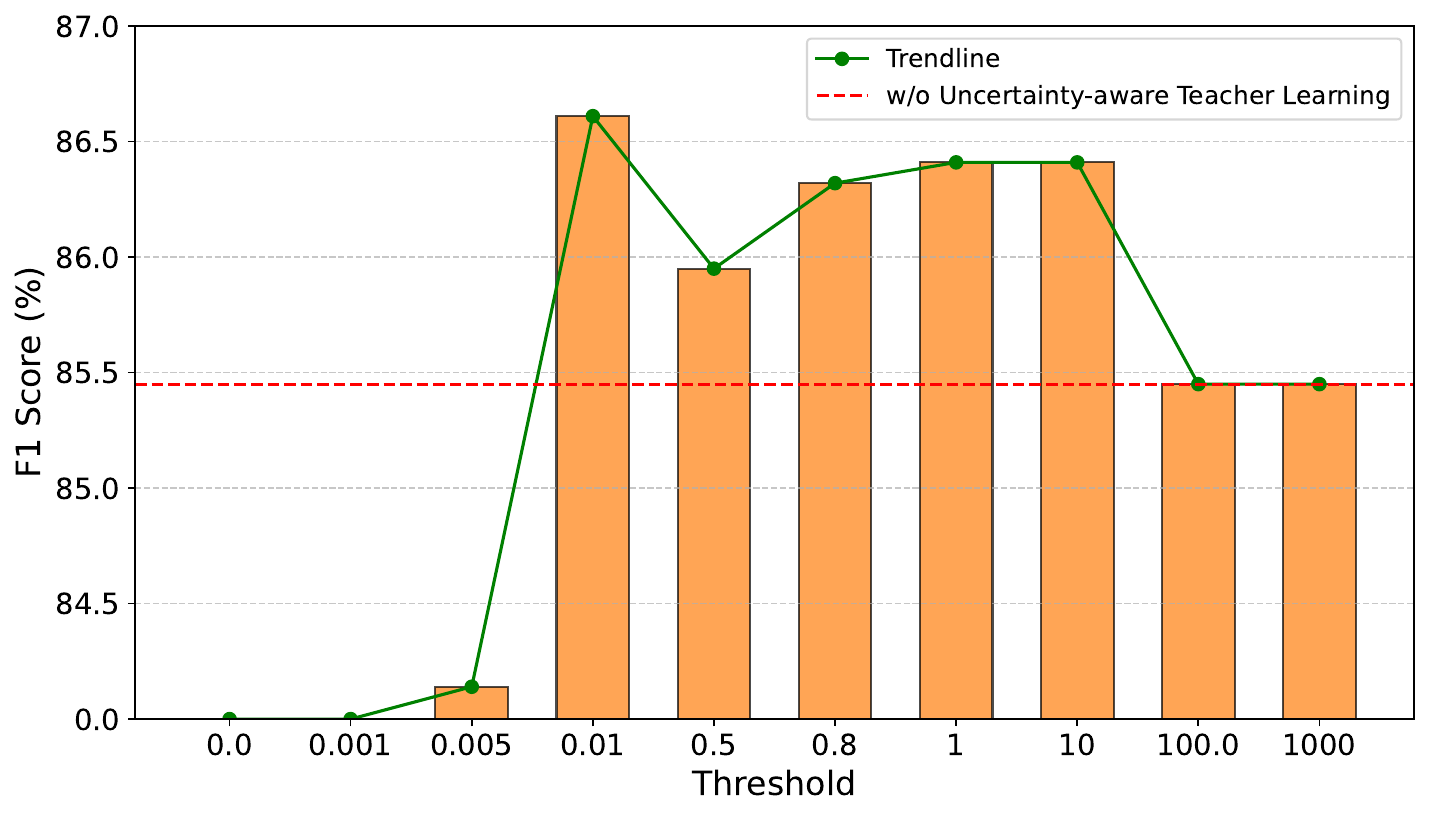}
    }
    \caption{F1 on CoNLL03 with different threshold $\sigma_{ua}$ in Uncertainty-Aware Label Selection.}
    \label{fig:UA}
\end{figure}

\begin{figure}
    \centering
    \resizebox{\linewidth}{!}{%
        \includegraphics[width=7.5cm]{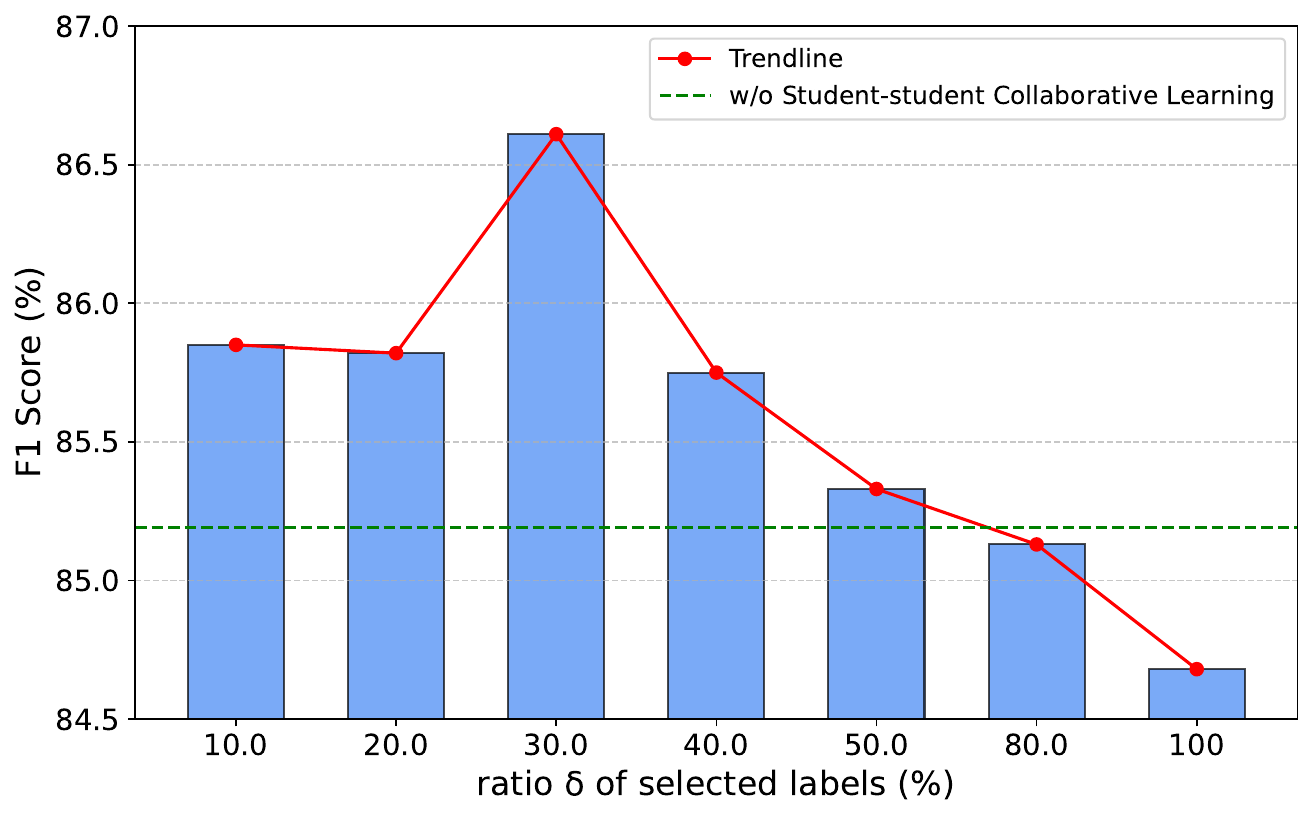}
    }
    \caption{F1 on CoNLL03 with different ratio $\delta$ of selected labels in Student-student Collaborative Learning.}
    \label{fig:Coteaching}
\end{figure}

\paragraph{Parameter Study}
As shown in Figure \ref{fig:UA} and Figure \ref{fig:Coteaching}, we conduct experiments to explore the impact of important hyperparameters to further understand Uncertainty-Aware Label Selection and Student-Student Collaborative Learning.
% Overall, 
% Overall, although the selection of different hyperparameters will have some impact on the model performance, when the hyperparameters are chosen reasonably rather than extreme values, the model always achieves an improvement in effectiveness compared to modules that do not use the design.
Overall, although the choice of different hyperparameters will have some impact on the model performance, as long as the hyperparameters are chosen wisely rather than at extreme values (e.g., wrongly setting the threshold $\sigma_{ua}$ in Uncertainty-Aware Label Selection to 0), the performance of the model will always be improved over what it would have been without using the components.
More detailed analysis are shown in the Appendix \ref{appendix_cparameter}.

\begin{table*}[h]
\small
\centering
\renewcommand{\arraystretch}{1.05}
\begin{tabular}{l}
% \hline
\toprule
\textbf{Distant Match}:  \textcolor[rgb]{1,0,0}{[Johnson]$_{\mathrm{PER}}$} is the second manager to be hospitalized after California \textcolor[rgb]{1,0,0}{[Angels]$_{\mathrm{PER}}$} \\skipper \textcolor[rgb]{1,0,0}{[John]$_{\mathrm{PER}}$} McNamara was admitted to New \textcolor[rgb]{1,0,0}{[York]$_{\mathrm{PER}}$} ’s \textcolor[rgb]{1,0,0}{[Columbia]$_{\mathrm{PER}}$} Presby Hospital .\\
\textbf{Ground Truth}:  \textcolor[rgb]{1,0,0}{[Johnson]$_{\mathrm{PER}}$} is the second manager to be hospitalized after \textcolor[rgb]{0,0,0.7}{[California Angels]$_{\mathrm{ORG}}$} \\skipper \textcolor[rgb]{1,0,0}{[John McNamara]$_{\mathrm{PER}}$} was admitted to \textcolor[rgb]{0,0.7,0}{[New York]$_{\mathrm{LOC}}$} ’s \textcolor[rgb]{0,0,0.7}{[Columbia Presby Hospital]$_{\mathrm{ORG}}$} .\\
\hline
\textbf{BOND}: \textcolor[rgb]{1,0,0}{[Johnson]$_{\mathrm{PER}}$} is the second manager to be hospitalized after \underline{\textcolor[rgb]{0,0.7,0}{[California]$_{\mathrm{LOC}}$} \textcolor[rgb]{1,0,0}{[Angels]$_{\mathrm{PER}}$}} \\skipper \textcolor[rgb]{1,0,0}{[John McNamara]$_{\mathrm{PER}}$} was admitted to \textcolor[rgb]{0,0.7,0}{[New York]$_{\mathrm{LOC}}$} ’s \textcolor[rgb]{1,0,0}{[Columbia]$_{\mathrm{PER}}$} Presby Hospital.\\

\textbf{SCDL}: \textcolor[rgb]{1,0,0}{[Johnson]$_{\mathrm{PER}}$} is the second manager to be hospitalized after \underline{\textcolor[rgb]{0,0.7,0}{[California]$_{\mathrm{LOC}}$} \textcolor[rgb]{1,0,0}{[Angels]$_{\mathrm{PER}}$}} \\skipper \textcolor[rgb]{1,0,0}{[John McNamara]$_{\mathrm{PER}}$} was admitted to \textcolor[rgb]{0,0.7,0}{[New York]$_{\mathrm{LOC}}$} ’s \textcolor[rgb]{0,0,0.7}{[Columbia Presby Hospital]$_{\mathrm{ORG}}$} .\\

\textbf{ATSEN}: \textcolor[rgb]{1,0,0}{[Johnson]$_{\mathrm{PER}}$} is the second manager to be hospitalized after \textcolor[rgb]{0,0,0.7}{[California Angels]$_{\mathrm{ORG}}$} \\skipper \textcolor[rgb]{1,0,0}{[John McNamara]$_{\mathrm{PER}}$} was admitted to \textcolor[rgb]{0,0.7,0}{[New York]$_{\mathrm{LOC}}$} ’s \textcolor[rgb]{0,0,0.7}{[Columbia Presby Hospital]$_{\mathrm{ORG}}$} .\\

\midrule
\textbf{CENSOR}: \textcolor[rgb]{1,0,0}{[Johnson]$_{\mathrm{PER}}$} is the second manager to be hospitalized after \textcolor[rgb]{0,0,0.7}{[California Angels]$_{\mathrm{ORG}}$} \\skipper \textcolor[rgb]{1,0,0}{[John McNamara]$_{\mathrm{PER}}$} was admitted to \textcolor[rgb]{0,0.7,0}{[New York]$_{\mathrm{LOC}}$} ’s \textcolor[rgb]{0,0,0.7}{[Columbia Presby Hospital]$_{\mathrm{ORG}}$} .\\
% \hline
\toprule
\end{tabular}
\caption{Case study with CENSOR and previous teacher-student methods for DS-NER. The sentence is from the CoNLL03 training set.}
\label{tab_case_1} 
\end{table*}

\begin{table*}[h]
\small
\centering
\renewcommand{\arraystretch}{1.05}
\begin{tabular}{l}
% \hline
\toprule

\textbf{Ground Truth}: All-conquering \textcolor[rgb]{0,0,0.7}{[Juventus]$_{\mathrm{ORG}}$} field their most recent signing, \textcolor[rgb]{1,0.4,0.7}{[Portuguese]$_{\mathrm{MISC}}$} defender \textcolor[rgb]{1,0,0}{[Dimas]$_{\mathrm{PER}}$}, \\ while \textcolor[rgb]{1,0,0}{[Alessandro Del Piero]$_{\mathrm{PER}}$} and \textcolor[rgb]{1,0.4,0.7}{[Croat]$_{\mathrm{MISC}}$} 
 \textcolor[rgb]{1,0,0}{[Alen Boksic]$_{\mathrm{PER}}$} lead the attack. \\

\hline
\textbf{BOND}: All-conquering \textcolor[rgb]{0,0,0.7}{[Juventus]$_{\mathrm{ORG}}$} field their most recent signing, \textcolor[rgb]{0,0,0.7}{[Portuguese]$_{\mathrm{ORG}}$} defender \textcolor[rgb]{1,0,0}{[Dimas]$_{\mathrm{PER}}$}, \\ while \textcolor[rgb]{1,0,0}{[Alessandro Del Piero]$_{\mathrm{PER}}$} and \textcolor[rgb]{1,0,0}{[Croat Alen Boksic]$_{\mathrm{PER}}$} lead the attack. \\
% \textbf{BOND}: All-conquering \textcolor[rgb]{0,0,0.7}{[Juventus]$_{\mathrm{ORG}}$} field their most recent signing, \textcolor[rgb]{1,0.4,0.7}{[Portuguese]$_{\mathrm{MISC}}$} defender \textcolor[rgb]{1,0,0}{[Dimas]$_{\mathrm{PER}}$}, \\ while \textcolor[rgb]{1,0,0}{[Alessandro Del Piero]$_{\mathrm{PER}}$} and \textcolor[rgb]{1,0,0}{[Croat Alen Boksic]$_{\mathrm{PER}}$} lead the attack. \\

\textbf{SCDL}: All-conquering \textcolor[rgb]{0,0,0.7}{[Juventus]$_{\mathrm{ORG}}$} field their most recent signing, \textcolor[rgb]{1,0.4,0.7}{[Portuguese]$_{\mathrm{MISC}}$} defender \textcolor[rgb]{1,0,0}{[Dimas]$_{\mathrm{PER}}$}, \\ while \textcolor[rgb]{1,0,0}{[Alessandro Del Piero]$_{\mathrm{PER}}$} and \textcolor[rgb]{1,0,0}{[Croat Alen Boksic]$_{\mathrm{PER}}$} lead the attack. \\

\textbf{ATSEN}: All-conquering \textcolor[rgb]{0,0,0.7}{[Juventus]$_{\mathrm{ORG}}$} field their most recent signing, \textcolor[rgb]{1,0.4,0.7}{[Portuguese]$_{\mathrm{MISC}}$} defender \textcolor[rgb]{1,0,0}{[Dimas]$_{\mathrm{PER}}$}, \\ while \textcolor[rgb]{1,0,0}{[Alessandro Del Piero]$_{\mathrm{PER}}$} and \textcolor[rgb]{0,0,0.7}{[Croat]$_{\mathrm{ORG}}$} 
 \textcolor[rgb]{1,0,0}{[Alen Boksic]$_{\mathrm{PER}}$} lead the attack. \\

\midrule
\textbf{CENSOR}: All-conquering \textcolor[rgb]{0,0,0.7}{[Juventus]$_{\mathrm{ORG}}$} field their most recent signing, \textcolor[rgb]{1,0.4,0.7}{[Portuguese]$_{\mathrm{MISC}}$} defender \textcolor[rgb]{1,0,0}{[Dimas]$_{\mathrm{PER}}$}, \\ while \textcolor[rgb]{1,0,0}{[Alessandro Del Piero]$_{\mathrm{PER}}$} and \textcolor[rgb]{1,0.4,0.7}{[Croat]$_{\mathrm{MISC}}$} 
 \textcolor[rgb]{1,0,0}{[Alen Boksic]$_{\mathrm{PER}}$} lead the attack. \\
 
% \hline
\bottomrule
\end{tabular}
\caption{Case study with CENSOR and previous teacher-student methods for DS-NER. The sentence is from the CoNLL03 test set.}
\label{tab_case_2} 
\end{table*}

\paragraph{Case Study}
% We also conduct a case study to show the advantages of the proposed CENSOR.
% % We compare 
% We show the prediction of several teacher-student methods for DS-NER, including BOND, SCDL, and ATSEN.
We also conduct the case study to understand the
advantage CENSOR with two examples
in Table \ref{tab_case_1} and Table \ref{tab_case_2}.
We show the prediction of BOND, SCDL, ATSEN and CENSOR on a training sequence with label noise and a test sequence with ground truth.
As shown in Table \ref{tab_case_1}, BOND and SCDL can slightly generalize to unseen mentions and relieve partial incomplete annotation, e.g., they can successfully recognize the ``John McNamara" and ``New York”.
However, these methods still suffer from label noise.
For comparison, for hard labels ``California Angels", CENSOR and ATSEN are able to detect them with advanced teacher-student design (e.g., Adaptive Teacher Learning in ATSEN and Student-Student Collaborative Learning in CENSOR) instead of relying purely on distant labels.
However, as shown in Table \ref{tab_case_2}, ATSEN still struggles to distinguish between easily confused samples and achieves inadequate generalization.
In contrast, as CENSOR can use fewer incorrect pseudo-labeled samples due to Uncertainty-Aware Teacher Learning and Student-Student Collaborative Learning, a higher degree of robustness and generalization can be achieved.

\section{Related Work}
\label{related_work}
\noindent
To alleviate the burden of annotation, previous studies attempted to annotate NER datasets via distant supervision, which suffers from noisy annotation.

\paragraph{DS-NER Methods}
To address these issues, various methods have been proposed.
Several studies \citep{shang-etal-2018-learning, DBLP:conf/coling/YangCLHZ18, jie-etal-2019-better} modify CRF to get better performance under the noise.
\citet{DBLP:conf/acl/PengXZFH19,zhou-etal-2022-distantly} try to employ PU learning to obtain the unbiased estimation of loss value. 
\citet{DBLP:conf/iclr/LiL021, li-etal-2022-rethinking} introduce negative sampling to mitigate the misguidance from unlabeled entities. 
\citet{DBLP:conf/kdd/LiangYJEWZZ20, zhang-etal-2021-improving, DBLP:conf/aaai/QuZLWHZ23} adopt the teacher-student framework to handle both inaccurate and incomplete labels simultaneously.
In this paper, we attempt to reduce the effect of incorrect pseudo labels and error propagation in the teacher-student framework to achieve better performance.

\paragraph{Teacher-Student Framework}
Teacher-student framework is a popular architecture in many semi-supervised tasks \citep{DBLP:conf/cvpr/HuoXHYZL021}.
Recently, the teacher-student framework has attracted increasing attention in DS-NER task.
BOND \citep{DBLP:conf/kdd/LiangYJEWZZ20} firstly attempts to apply self-training with a teacher-student network in DS-NER.
SCDL \citep{zhang-etal-2021-improving} further improves the performance by jointly training two teacher-student networks.
ATSEN \citep{DBLP:conf/aaai/QuZLWHZ23} considers both consistent and inconsistent predictions between two teachers and proposes fine-grained teacher updating to achieve more robustness.
We improve the teacher-student framework by Uncertainty-Aware Teacher Learning and Student-Student Collaborative Learning, jointly reducing the effect of incorrect pseudo labels.
In this way, our method can avoid error propagation and achieve better overall performance.

\section{Conclusion}
% In this paper, we present a novel self-training framework
% NINTENDO for DS-NER. 
% Specifically, NINTENDO adopts Uncertainty-aware Teacher Learning that leverages the prediction uncertainty to guide the selection procedure of pseudo labels, reducing the number of incorrect pseudo labels generated by relying on confidence scores from poorly calibrated teacher networks.
% Furthermore, we devise Student-student Collaborative Learning to allow the transfer of reliable labels between two. 
% With it, a student network is able to not indiscriminately rely on all pseudo labels from its corresponding teacher network and further reduce the possibility of learning incorrect pseudo labels from the teacher network.
% The experiment results illustrate that NINTENDO significantly outperforms previous methods.
In this paper, we introduce CENSOR, a novel teacher-student framework designed for the DS-NER task. 
CENSOR firstly incorporates Uncertainty-Aware Teacher Learning (UTL), utilizing prediction uncertainty to guide the pseudo-label selection. 
It mitigates the usage of incorrect pseudo labels by avoiding reliance on confidence scores from poorly calibrated teacher networks.
We also introduce Student-Student Collaborative Learning (SCL) to enable a student network not to completely rely on pseudo labels from its teacher network, minimizing the risk of learning incorrect ones.
% This component also ensures that the training set is comprehensively utilized.
Meanwhile, this component component also ensures that the training set is comprehensively utilized.
% Our experimental results demonstrate CENSOR's superior performance compared to previous methods.
Our experimental results substantiate that CENSOR achieves superior performance compared to prior methodologies.

% Our code will be open-sourced after peer review.
% Inaccurate and incomplete annotation noise are two types
% of noise in DS-NER. 
% We propose SANTA to use separate strategies for two types of noise.
% For inaccurate annotation, we propose Memory-smoothed Focal Loss and Entity-aware KNN to relief the ambiguity problem. 
% For incomplete annotation, we utilize noise-robust loss GCE + SR and propose Boundary Mixup to improve the robustness and mitigate the decision boundary shifting problem.
% Experiments show that SANTA achieves state-of-the-art methods on five DS-NER datasets and the separate strategies are effective.

\section*{Limitations}
Our proposed CENSOR has two tiny limitations, specifically: (1) CENSOR focuses on addressing the label noise in the DS-NER task, and all our analyses are specific to this task. 
As a result, our model may not be robust enough compared to other models if it is not specific to the DS-NER task.
(2) Due to introducing the proposed Uncertainty-Aware Teacher Learning, our model will perform multiple forward passes in the uncertainty estimation phase, increasing the self-training time.
Compared to ATSEN, the self-training of our model takes about 4 times as long as that of ATSEN.
We plan to address these shortcomings as future work.

\section*{Acknowledgements}
This work is supported by the National Science Foundation of China under Grant No.61936012 and 61876004.
Meanwhile, we would like to thank the anonymous reviewers for their thoughtful and constructive comments.
Our code will be available at \href{https://github.com/PKUnlp-icler/CENSOR}{https://github.com/PKUnlp-icler/CENSOR}.

\bibliography{anthology,custom}

\begin{thebibliography}{36}
\providecommand{\natexlab}[1]{#1}

\bibitem[{Arpit et~al.(2017)Arpit, Jastrzebski, Ballas, Krueger, Bengio, Kanwal, Maharaj, Fischer, Courville, Bengio, and Lacoste{-}Julien}]{DBLP:conf/icml/ArpitJBKBKMFCBL17}
Devansh Arpit, Stanislaw Jastrzebski, Nicolas Ballas, David Krueger, Emmanuel Bengio, Maxinder~S. Kanwal, Tegan Maharaj, Asja Fischer, Aaron~C. Courville, Yoshua Bengio, and Simon Lacoste{-}Julien. 2017.
\newblock \href {http://proceedings.mlr.press/v70/arpit17a.html} {A closer look at memorization in deep networks}.
\newblock In \emph{Proceedings of the 34th International Conference on Machine Learning, {ICML} 2017, Sydney, NSW, Australia, 6-11 August 2017}, volume~70 of \emph{Proceedings of Machine Learning Research}, pages 233--242. {PMLR}.

\bibitem[{Balasuriya et~al.(2009)Balasuriya, Ringland, Nothman, Murphy, and Curran}]{DBLP:conf/acl-pwnlp/BalasuriyaRNMC09}
Dominic Balasuriya, Nicky Ringland, Joel Nothman, Tara Murphy, and James~R. Curran. 2009.
\newblock \href {https://aclanthology.org/W09-3302/} {Named entity recognition in wikipedia}.
\newblock In \emph{Proceedings of the 1st 2009 Workshop on The People's Web Meets {NLP:} Collaboratively Constructed Semantic Resources@IJCNLP 2009, Suntec, Singapore, August 7, 2009}, pages 10--18. Association for Computational Linguistics.

\bibitem[{Cao et~al.(2019)Cao, Hu, Chua, Liu, and Ji}]{DBLP:conf/emnlp/CaoHCLJ19}
Yixin Cao, Zikun Hu, Tat{-}Seng Chua, Zhiyuan Liu, and Heng Ji. 2019.
\newblock \href {https://doi.org/10.18653/v1/D19-1025} {Low-resource name tagging learned with weakly labeled data}.
\newblock In \emph{Proceedings of the 2019 Conference on Empirical Methods in Natural Language Processing and the 9th International Joint Conference on Natural Language Processing, {EMNLP-IJCNLP} 2019, Hong Kong, China, November 3-7, 2019}, pages 261--270. Association for Computational Linguistics.

\bibitem[{Feng et~al.(2019)Feng, Tao, Wu, Feng, Zhao, and Yan}]{feng-etal-2019-learning}
Jiazhan Feng, Chongyang Tao, Wei Wu, Yansong Feng, Dongyan Zhao, and Rui Yan. 2019.
\newblock \href {https://doi.org/10.18653/v1/P19-1370} {Learning a matching model with co-teaching for multi-turn response selection in retrieval-based dialogue systems}.
\newblock In \emph{Proceedings of the 57th Annual Meeting of the Association for Computational Linguistics}, pages 3805--3815, Florence, Italy. Association for Computational Linguistics.

\bibitem[{Godin et~al.(2015)Godin, Vandersmissen, Neve, and de~Walle}]{DBLP:conf/aclnut/GodinVNW15}
Fr{\'{e}}deric Godin, Baptist Vandersmissen, Wesley~De Neve, and Rik~Van de~Walle. 2015.
\newblock \href {https://doi.org/10.18653/V1/W15-4322} {Multimedia lab {\textdollar}@{\textdollar} {ACL} {WNUT} {NER} shared task: Named entity recognition for twitter microposts using distributed word representations}.
\newblock In \emph{Proceedings of the Workshop on Noisy User-generated Text, NUT@IJCNLP 2015, Beijing, China, July 31, 2015}, pages 146--153. Association for Computational Linguistics.

\bibitem[{Guo et~al.(2017)Guo, Pleiss, Sun, and Weinberger}]{DBLP:conf/icml/GuoPSW17}
Chuan Guo, Geoff Pleiss, Yu~Sun, and Kilian~Q. Weinberger. 2017.
\newblock \href {http://proceedings.mlr.press/v70/guo17a.html} {On calibration of modern neural networks}.
\newblock In \emph{Proceedings of the 34th International Conference on Machine Learning, {ICML} 2017, Sydney, NSW, Australia, 6-11 August 2017}, volume~70 of \emph{Proceedings of Machine Learning Research}, pages 1321--1330. {PMLR}.

\bibitem[{Han et~al.(2018)Han, Yao, Yu, Niu, Xu, Hu, Tsang, and Sugiyama}]{DBLP:conf/nips/HanYYNXHTS18}
Bo~Han, Quanming Yao, Xingrui Yu, Gang Niu, Miao Xu, Weihua Hu, Ivor~W. Tsang, and Masashi Sugiyama. 2018.
\newblock \href {https://proceedings.neurips.cc/paper/2018/hash/a19744e268754fb0148b017647355b7b-Abstract.html} {Co-teaching: Robust training of deep neural networks with extremely noisy labels}.
\newblock In \emph{Advances in Neural Information Processing Systems 31: Annual Conference on Neural Information Processing Systems 2018, NeurIPS 2018, December 3-8, 2018, Montr{\'{e}}al, Canada}, pages 8536--8546.

\bibitem[{Huo et~al.(2021)Huo, Xie, He, Yang, Zhou, Li, and Tian}]{DBLP:conf/cvpr/HuoXHYZL021}
Xinyue Huo, Lingxi Xie, Jianzhong He, Zijie Yang, Wengang Zhou, Houqiang Li, and Qi~Tian. 2021.
\newblock \href {https://doi.org/10.1109/CVPR46437.2021.00129} {{ATSO:} asynchronous teacher-student optimization for semi-supervised image segmentation}.
\newblock In \emph{{IEEE} Conference on Computer Vision and Pattern Recognition, {CVPR} 2021, virtual, June 19-25, 2021}, pages 1235--1244. Computer Vision Foundation / {IEEE}.

\bibitem[{Jie et~al.(2019)Jie, Xie, Lu, Ding, and Li}]{jie-etal-2019-better}
Zhanming Jie, Pengjun Xie, Wei Lu, Ruixue Ding, and Linlin Li. 2019.
\newblock \href {https://doi.org/10.18653/v1/N19-1079} {Better modeling of incomplete annotations for named entity recognition}.
\newblock In \emph{Proceedings of the 2019 Conference of the North {A}merican Chapter of the Association for Computational Linguistics: Human Language Technologies, Volume 1 (Long and Short Papers)}, pages 729--734, Minneapolis, Minnesota. Association for Computational Linguistics.

\bibitem[{Kingma and Ba(2015)}]{DBLP:journals/corr/KingmaB14}
Diederik~P. Kingma and Jimmy Ba. 2015.
\newblock \href {http://arxiv.org/abs/1412.6980} {Adam: {A} method for stochastic optimization}.
\newblock In \emph{3rd International Conference on Learning Representations, {ICLR} 2015, San Diego, CA, USA, May 7-9, 2015, Conference Track Proceedings}.

\bibitem[{Krizhevsky et~al.(2012)Krizhevsky, Sutskever, and Hinton}]{DBLP:conf/nips/KrizhevskySH12}
Alex Krizhevsky, Ilya Sutskever, and Geoffrey~E. Hinton. 2012.
\newblock \href {https://proceedings.neurips.cc/paper/2012/hash/c399862d3b9d6b76c8436e924a68c45b-Abstract.html} {Imagenet classification with deep convolutional neural networks}.
\newblock In \emph{Advances in Neural Information Processing Systems 25: 26th Annual Conference on Neural Information Processing Systems 2012. Proceedings of a meeting held December 3-6, 2012, Lake Tahoe, Nevada, United States}, pages 1106--1114.

\bibitem[{Li et~al.(2021)Li, Liu, and Shi}]{DBLP:conf/iclr/LiL021}
Yangming Li, Lemao Liu, and Shuming Shi. 2021.
\newblock \href {https://openreview.net/forum?id=5jRVa89sZk} {Empirical analysis of unlabeled entity problem in named entity recognition}.
\newblock In \emph{9th International Conference on Learning Representations, {ICLR} 2021, Virtual Event, Austria, May 3-7, 2021}. OpenReview.net.

\bibitem[{Li et~al.(2022)Li, Liu, and Shi}]{li-etal-2022-rethinking}
Yangming Li, Lemao Liu, and Shuming Shi. 2022.
\newblock \href {https://doi.org/10.18653/v1/2022.acl-long.497} {Rethinking negative sampling for handling missing entity annotations}.
\newblock In \emph{Proceedings of the 60th Annual Meeting of the Association for Computational Linguistics (Volume 1: Long Papers)}, pages 7188--7197, Dublin, Ireland. Association for Computational Linguistics.

\bibitem[{Li and Zhao(2023)}]{li-zhao-2023-em}
Yiyang Li and Hai Zhao. 2023.
\newblock \href {https://doi.org/10.18653/v1/2023.acl-long.7} {{EM} pre-training for multi-party dialogue response generation}.
\newblock In \emph{Proceedings of the 61st Annual Meeting of the Association for Computational Linguistics (Volume 1: Long Papers)}, pages 92--103, Toronto, Canada. Association for Computational Linguistics.

\bibitem[{Liang et~al.(2020)Liang, Yu, Jiang, Er, Wang, Zhao, and Zhang}]{DBLP:conf/kdd/LiangYJEWZZ20}
Chen Liang, Yue Yu, Haoming Jiang, Siawpeng Er, Ruijia Wang, Tuo Zhao, and Chao Zhang. 2020.
\newblock \href {https://doi.org/10.1145/3394486.3403149} {{BOND:} bert-assisted open-domain named entity recognition with distant supervision}.
\newblock In \emph{{KDD} '20: The 26th {ACM} {SIGKDD} Conference on Knowledge Discovery and Data Mining, Virtual Event, CA, USA, August 23-27, 2020}, pages 1054--1064. {ACM}.

\bibitem[{Liu et~al.(2023)Liu, Jiang, Yin, Wang, Mi, Liu, Wan, and Wang}]{liu-etal-2023-one}
Yajiao Liu, Xin Jiang, Yichun Yin, Yasheng Wang, Fei Mi, Qun Liu, Xiang Wan, and Benyou Wang. 2023.
\newblock \href {https://doi.org/10.18653/v1/2023.acl-long.1} {One cannot stand for everyone! leveraging multiple user simulators to train task-oriented dialogue systems}.
\newblock In \emph{Proceedings of the 61st Annual Meeting of the Association for Computational Linguistics (Volume 1: Long Papers)}, pages 1--21, Toronto, Canada. Association for Computational Linguistics.

\bibitem[{Liu et~al.(2019)Liu, Ott, Goyal, Du, Joshi, Chen, Levy, Lewis, Zettlemoyer, and Stoyanov}]{liu2019roberta}
Yinhan Liu, Myle Ott, Naman Goyal, Jingfei Du, Mandar Joshi, Danqi Chen, Omer Levy, Mike Lewis, Luke Zettlemoyer, and Veselin Stoyanov. 2019.
\newblock Roberta: A robustly optimized bert pretraining approach.
\newblock \emph{arXiv preprint arXiv:1907.11692}.

\bibitem[{Ma and Hovy(2016)}]{ma-hovy-2016-end}
Xuezhe Ma and Eduard Hovy. 2016.
\newblock \href {https://doi.org/10.18653/v1/P16-1101} {End-to-end sequence labeling via bi-directional {LSTM}-{CNN}s-{CRF}}.
\newblock In \emph{Proceedings of the 54th Annual Meeting of the Association for Computational Linguistics (Volume 1: Long Papers)}, pages 1064--1074, Berlin, Germany. Association for Computational Linguistics.

\bibitem[{Peng et~al.(2019)Peng, Xing, Zhang, Fu, and Huang}]{DBLP:conf/acl/PengXZFH19}
Minlong Peng, Xiaoyu Xing, Qi~Zhang, Jinlan Fu, and Xuanjing Huang. 2019.
\newblock \href {https://doi.org/10.18653/v1/p19-1231} {Distantly supervised named entity recognition using positive-unlabeled learning}.
\newblock In \emph{Proceedings of the 57th Conference of the Association for Computational Linguistics, {ACL} 2019, Florence, Italy, July 28- August 2, 2019, Volume 1: Long Papers}, pages 2409--2419. Association for Computational Linguistics.

\bibitem[{Qu et~al.(2023)Qu, Zeng, Liu, Wang, Huai, and Zhou}]{DBLP:conf/aaai/QuZLWHZ23}
Xiaoye Qu, Jun Zeng, Daizong Liu, Zhefeng Wang, Baoxing Huai, and Pan Zhou. 2023.
\newblock \href {https://doi.org/10.1609/AAAI.V37I11.26583} {Distantly-supervised named entity recognition with adaptive teacher learning and fine-grained student ensemble}.
\newblock In \emph{Thirty-Seventh {AAAI} Conference on Artificial Intelligence, {AAAI} 2023, Thirty-Fifth Conference on Innovative Applications of Artificial Intelligence, {IAAI} 2023, Thirteenth Symposium on Educational Advances in Artificial Intelligence, {EAAI} 2023, Washington, DC, USA, February 7-14, 2023}, pages 13501--13509. {AAAI} Press.

\bibitem[{Ratinov and Roth(2009)}]{ratinov-roth-2009-design}
Lev Ratinov and Dan Roth. 2009.
\newblock \href {https://aclanthology.org/W09-1119} {Design challenges and misconceptions in named entity recognition}.
\newblock In \emph{Proceedings of the Thirteenth Conference on Computational Natural Language Learning ({C}o{NLL}-2009)}, pages 147--155, Boulder, Colorado. Association for Computational Linguistics.

\bibitem[{Rizve et~al.(2021)Rizve, Duarte, Rawat, and Shah}]{DBLP:conf/iclr/RizveDRS21}
Mamshad~Nayeem Rizve, Kevin Duarte, Yogesh~S. Rawat, and Mubarak Shah. 2021.
\newblock \href {https://openreview.net/forum?id=-ODN6SbiUU} {In defense of pseudo-labeling: An uncertainty-aware pseudo-label selection framework for semi-supervised learning}.
\newblock In \emph{9th International Conference on Learning Representations, {ICLR} 2021, Virtual Event, Austria, May 3-7, 2021}. OpenReview.net.

\bibitem[{Sanh et~al.(2019)Sanh, Debut, Chaumond, and Wolf}]{Sanh2019DistilBERTAD}
Victor Sanh, Lysandre Debut, Julien Chaumond, and Thomas Wolf. 2019.
\newblock \href {https://api.semanticscholar.org/CorpusID:203626972} {Distilbert, a distilled version of bert: smaller, faster, cheaper and lighter}.
\newblock \emph{ArXiv}, abs/1910.01108.

\bibitem[{Shang et~al.(2018)Shang, Liu, Gu, Ren, Ren, and Han}]{shang-etal-2018-learning}
Jingbo Shang, Liyuan Liu, Xiaotao Gu, Xiang Ren, Teng Ren, and Jiawei Han. 2018.
\newblock \href {https://doi.org/10.18653/v1/D18-1230} {Learning named entity tagger using domain-specific dictionary}.
\newblock In \emph{Proceedings of the 2018 Conference on Empirical Methods in Natural Language Processing}, pages 2054--2064, Brussels, Belgium. Association for Computational Linguistics.

\bibitem[{Si et~al.(2023)Si, Cai, Zeng, Feng, Lin, and Chang}]{si-etal-2023-santa}
Shuzheng Si, Zefan Cai, Shuang Zeng, Guoqiang Feng, Jiaxing Lin, and Baobao Chang. 2023.
\newblock \href {https://doi.org/10.18653/v1/2023.findings-acl.239} {{SANTA}: Separate strategies for inaccurate and incomplete annotation noise in distantly-supervised named entity recognition}.
\newblock In \emph{Findings of the Association for Computational Linguistics: ACL 2023}, pages 3883--3896, Toronto, Canada. Association for Computational Linguistics.

\bibitem[{Si et~al.(2024)Si, Ma, Gao, Wu, Lin, Dai, Li, Yan, Huang, and Li}]{si2024spokenwoz}
Shuzheng Si, Wentao Ma, Haoyu Gao, Yuchuan Wu, Ting-En Lin, Yinpei Dai, Hangyu Li, Rui Yan, Fei Huang, and Yongbin Li. 2024.
\newblock Spokenwoz: A large-scale speech-text benchmark for spoken task-oriented dialogue agents.
\newblock \emph{Advances in Neural Information Processing Systems}, 36.

\bibitem[{Si et~al.(2022{\natexlab{a}})Si, Zeng, and Chang}]{si-etal-2022-mining}
Shuzheng Si, Shuang Zeng, and Baobao Chang. 2022{\natexlab{a}}.
\newblock \href {https://doi.org/10.18653/v1/2022.naacl-main.356} {Mining clues from incomplete utterance: A query-enhanced network for incomplete utterance rewriting}.
\newblock In \emph{Proceedings of the 2022 Conference of the North American Chapter of the Association for Computational Linguistics: Human Language Technologies}, pages 4839--4847, Seattle, United States. Association for Computational Linguistics.

\bibitem[{Si et~al.(2022{\natexlab{b}})Si, Zeng, Lin, and Chang}]{si-etal-2022-scl}
Shuzheng Si, Shuang Zeng, Jiaxing Lin, and Baobao Chang. 2022{\natexlab{b}}.
\newblock \href {https://aclanthology.org/2022.coling-1.202} {{SCL}-{RAI}: Span-based contrastive learning with retrieval augmented inference for unlabeled entity problem in {NER}}.
\newblock In \emph{Proceedings of the 29th International Conference on Computational Linguistics}, pages 2313--2318, Gyeongju, Republic of Korea. International Committee on Computational Linguistics.

\bibitem[{Tjong Kim~Sang and De~Meulder(2003)}]{tjong-kim-sang-de-meulder-2003-introduction}
Erik~F. Tjong Kim~Sang and Fien De~Meulder. 2003.
\newblock \href {https://aclanthology.org/W03-0419} {Introduction to the {C}o{NLL}-2003 shared task: Language-independent named entity recognition}.
\newblock In \emph{Proceedings of the Seventh Conference on Natural Language Learning at {HLT}-{NAACL} 2003}, pages 142--147.

\bibitem[{Wei et~al.(2020)Wei, Feng, Chen, and An}]{DBLP:conf/cvpr/WeiFC020}
Hongxin Wei, Lei Feng, Xiangyu Chen, and Bo~An. 2020.
\newblock \href {https://doi.org/10.1109/CVPR42600.2020.01374} {Combating noisy labels by agreement: {A} joint training method with co-regularization}.
\newblock In \emph{2020 {IEEE/CVF} Conference on Computer Vision and Pattern Recognition, {CVPR} 2020, Seattle, WA, USA, June 13-19, 2020}, pages 13723--13732. Computer Vision Foundation / {IEEE}.

\bibitem[{Weischedel et~al.(2013)Weischedel, Palmer, Marcus, Hovy, Pradhan, Ramshaw, Xue, Taylor, Kaufman, Franchini et~al.}]{weischedel2013ontonotes}
Ralph Weischedel, Martha Palmer, Mitchell Marcus, Eduard Hovy, Sameer Pradhan, Lance Ramshaw, Nianwen Xue, Ann Taylor, Jeff Kaufman, Michelle Franchini, et~al. 2013.
\newblock Ontonotes release 5.0 ldc2013t19.
\newblock \emph{Linguistic Data Consortium, Philadelphia, PA}, 23.

\bibitem[{Yang et~al.(2018)Yang, Chen, Li, He, and Zhang}]{DBLP:conf/coling/YangCLHZ18}
YaoSheng Yang, Wenliang Chen, Zhenghua Li, Zhengqiu He, and Min Zhang. 2018.
\newblock \href {https://aclanthology.org/C18-1183/} {Distantly supervised {NER} with partial annotation learning and reinforcement learning}.
\newblock In \emph{Proceedings of the 27th International Conference on Computational Linguistics, {COLING} 2018, Santa Fe, New Mexico, USA, August 20-26, 2018}, pages 2159--2169. Association for Computational Linguistics.

\bibitem[{Yu et~al.(2019)Yu, Han, Yao, Niu, Tsang, and Sugiyama}]{DBLP:conf/icml/Yu0YNTS19}
Xingrui Yu, Bo~Han, Jiangchao Yao, Gang Niu, Ivor~W. Tsang, and Masashi Sugiyama. 2019.
\newblock \href {http://proceedings.mlr.press/v97/yu19b.html} {How does disagreement help generalization against label corruption?}
\newblock In \emph{Proceedings of the 36th International Conference on Machine Learning, {ICML} 2019, 9-15 June 2019, Long Beach, California, {USA}}, volume~97 of \emph{Proceedings of Machine Learning Research}, pages 7164--7173. {PMLR}.

\bibitem[{Zhang et~al.(2021{\natexlab{a}})Zhang, Yu, Liu, Zhang, Sheng, Mengge, and Xu}]{zhang-etal-2021-improving-distantly}
Xinghua Zhang, Bowen Yu, Tingwen Liu, Zhenyu Zhang, Jiawei Sheng, Xue Mengge, and Hongbo Xu. 2021{\natexlab{a}}.
\newblock \href {https://doi.org/10.18653/v1/2021.findings-emnlp.131} {Improving distantly-supervised named entity recognition with self-collaborative denoising learning}.
\newblock In \emph{Findings of the Association for Computational Linguistics: EMNLP 2021}, pages 1518--1529, Punta Cana, Dominican Republic. Association for Computational Linguistics.

\bibitem[{Zhang et~al.(2021{\natexlab{b}})Zhang, Yu, Shu, Mengge, Liu, and Guo}]{zhang-etal-2021-improving}
Zhenyu Zhang, Bowen Yu, Xiaobo Shu, Xue Mengge, Tingwen Liu, and Li~Guo. 2021{\natexlab{b}}.
\newblock \href {https://doi.org/10.18653/v1/2021.findings-acl.8} {From what to why: Improving relation extraction with rationale graph}.
\newblock In \emph{Findings of the Association for Computational Linguistics: ACL-IJCNLP 2021}, pages 86--95, Online. Association for Computational Linguistics.

\bibitem[{Zhou et~al.(2022)Zhou, Li, and Li}]{zhou-etal-2022-distantly}
Kang Zhou, Yuepei Li, and Qi~Li. 2022.
\newblock \href {https://doi.org/10.18653/v1/2022.acl-long.498} {Distantly supervised named entity recognition via confidence-based multi-class positive and unlabeled learning}.
\newblock In \emph{Proceedings of the 60th Annual Meeting of the Association for Computational Linguistics (Volume 1: Long Papers)}, pages 7198--7211, Dublin, Ireland. Association for Computational Linguistics.

\end{thebibliography}

\appendix

% \newpage
\section*{Appendix}

\section{DS-NER Datasets}
\label{appendix_dataset}
Statistics of five datasets are shown in Table \ref{tab_datasets}.

\begin{table}[h]
\renewcommand\arraystretch{1.1}
\centering
\scriptsize
% \resizebox{6.8cm}{2.2cm}{
\begin{tabular}{cccccc}
\toprule
\multicolumn{2}{c}{\textbf{Dataset}} & Train  & Dev  & Test  & Types \\ \midrule
\multirow{2}{*}{\textbf{CoNLL03}} & \textbf{Sentence}     & 14041      & 3250   & 3453  & \multirow{2}{*}{4}  \\
& \textbf{Token}            & 203621  & 51362  & 46435 \\ \hline
\multirow{2}{*}{\textbf{OntoNotes5.0}} & \textbf{Sentence}     & 115812     & 15680  & 12217  & \multirow{2}{*}{18}  \\
& \textbf{Token}            & 2200865 & 304701 & 230118 \\ \hline
\multirow{2}{*}{\textbf{Webpage}} & \textbf{Sentence}     & 385     & 99  & 135  & \multirow{2}{*}{4}  \\
& \textbf{Token}            & 5293 & 1121 & 1131 \\ \hline
\multirow{2}{*}{\textbf{Wikigold}} & \textbf{Sentence}     & 1142     & 280  & 274  & \multirow{2}{*}{4}  \\
& \textbf{Token}            & 25819 & 6650 & 6538 \\ \hline
\multirow{2}{*}{\textbf{Twitter}} & \textbf{Sentence}     & 2393     & 999  & 3844  & \multirow{2}{*}{10}  \\
& \textbf{Token}            & 44076 & 15262 & 58064 \\ \bottomrule
\end{tabular}%}
\caption{The statistics of five DS-NER datasets.}
\label{tab_datasets}
\end{table}

\section{Hyperparameters}
\label{appendix_hyperparameter}
Detailed hyperparameters are shown in Table \ref{tab_hyper}.
% We use Adam \citep{DBLP:journals/corr/KingmaB14} as our optimizer.
Experiments are run on a single NVIDIA A40.

\begin{algorithm}[h]
% \fontsize{6pt}{1pt}
% \footnotesize
\scriptsize
\setstretch{1.15}
\captionsetup{font={small}}
\caption{Training Procedure of CENSOR.}
\label{alg:algorithm}
\textbf{Input}: DS-NER dataset $D_{ds}=\{(X_i, Y_i)\}_{i=1}^N$ \\
\textbf{Parameter}: Two teacher-student network parameters, including $W_{t_1}$, $W_{s_1}$, $W_{t_2}$, and $W_{s_2}$ \\
% \textbf{Parameter}: Two network parameters $W_{A}$, $W_{B}$ \\
\textbf{Output}: The best model
\begin{algorithmic}[1] %[1] enables line numbers
\STATE Pre-training two models $W_A$, $W_B$ with $D_{ds}$. \hfill $\triangleright${\it Pre-Training}.\\
\STATE Initialize two teacher-student networks: $W_{t_1}\gets W_A$, $W_{s_1} \gets W_A$, $W_{t_2} \gets W_B$, $W_{s_2} \gets W_B$.\\
\STATE Initialize training step: $step \gets 0$.
% \FOR {$p \gets 1\ to\ P$}
\STATE Initialize noisy labels: $Y_{\uppercase\expandafter{\romannumeral1}} \gets Y, Y_{\uppercase\expandafter{\romannumeral2}} \gets Y$.\\
\WHILE{ \emph{not reach max training epochs} }
    \STATE Get a batch $\hat{D} = (X^{(b)}, Y^{(b)}_{\uppercase\expandafter{\romannumeral1}}, Y^{(b)}_{\uppercase\expandafter{\romannumeral2}})$ from $D_{ds}$, \\
    $step \gets step+1$. \hfill $\triangleright${\it Self-Training}.\\
    \STATE Get pseudo labels via the teacher $W_{t_1}$, $W_{t_2}$: \\ 
    \quad \quad$ \tilde{Y}^{(b)}_{\uppercase\expandafter{\romannumeral1}} \gets f(X^{(b)}; W_{t_1}) $, \\ 
    \quad \quad$ \tilde{Y}^{(b)}_{\uppercase\expandafter{\romannumeral2}} \gets f(X^{(b)}; W_{t_2})$. \\
    % \STATE Get masked matrix M by Eq.3, Eq.4 and Eq.5 \\ 
    \STATE Select reliable labels via Uncertainty-Aware Teacher Learning: \\
    \quad \quad Estimate Confidence and Uncertainty by Eq.3 and Eq.4, separately\\
    \quad \quad $\mathcal{T}^{(b)}_{\uppercase\expandafter{\romannumeral1}} \gets$ Uncertainty-Aware Label Selection$(Y^{(b)}_{\uppercase\expandafter{\romannumeral1}},  \tilde{Y}^{(b)}_{\uppercase\expandafter{\romannumeral1}})$, \\ \quad \quad $\mathcal{T}^{(b)}_{\uppercase\expandafter{\romannumeral2}} \gets$ Uncertainty-Aware Label Selection$(Y^{(b)}_{\uppercase\expandafter{\romannumeral2}},  \tilde{Y}^{(b)}_{\uppercase\expandafter{\romannumeral2}})$. \\
    \STATE Select reliable labels via Student-Student Collaborative Learning: \\ 
    \quad \quad$
    \hat{D}^*_{s_1} = \arg\min_{\hat{D}:|{\hat{D}}|\ge \delta\%|\hat{D}|}Loss(s_1, \hat{D})$, \\
    \hfill //sample $\delta\%$ small-loss instances \\ 
    \quad \quad$
    \hat{D}^*_{s_2} = \arg\min_{{\hat{D}}:|{\hat{D}}|\ge \delta\% |\hat{D}|}Loss(s_2, \hat{D})$. \\
    \hfill //sample $\delta\%$ small-loss instances \\
    % \\
    \quad \quad Transfer the pseudo labels between $\hat{D}^*_{s_1}$ and $\hat{D}^*_{s_2}$ . \\
    \STATE Update the student $W_{s_1}$ and $W_{s_2}$ by Eq. 7. \\
    \STATE Update the teacher $W_{t_1}$ and $W_{t_2}$ by Eq. 8. \\
    % \IF { $step \text{ mod } Update\_Cycle = 0$ }
    %     \STATE Update noisy labels mutually: \\
    %     \quad \quad $Y_{\uppercase\expandafter{\romannumeral1}}=\{Y_i \gets f(X_i; W_{t_1})\}_{i=1}^N$, \\
    %     \quad \quad $Y_{\uppercase\expandafter{\romannumeral2}}=\{Y_i \gets f(X_i; W_{t_2})\}_{i=1}^N$.
    % \ENDIF
\ENDWHILE
\STATE Evaluate models $W_{t_1}$, $W_{s_1}$, $W_{t_2}$, $W_{s_2}$ on {\it Dev} set. \\
\STATE \textbf{return} The best model $W\in\{ W_{t_1}, W_{s_1}, W_{t_2}, W_{s_2}\}$
\end{algorithmic}
\label{tab_code}
\end{algorithm}

\section{Pseudocode}
\label{appendix_pseudocode}
Algorithm \ref{tab_code} gives the pseudocode of our method.

\section{Robustness to Different Noise Ratios}
Detailed data in Figure \ref{fig:noise} can be found in Table \ref{tab_noise}.

\begin{table}[t]
% \centering
\scriptsize
\setlength\tabcolsep{3pt}
    \centering
        \begin{tabular}{cccccc}
    \toprule
    \textbf{Name} & \textbf{CoNLL03} & \textbf{Ont5.0} & \textbf{Webpage} & \textbf{Wikigold} & \textbf{Twitter}\\
    \midrule
    \textbf{Learning Rate} &{1e-5} &{2e-5} &{1e-5} &{1e-5} &{2e-5}  \\
    \midrule
    \textbf{Batch Size} &{8} &{16} &{16} &{16} &{8} \\
    \midrule
    \textbf{EMA} $\alpha$ &{0.995} &{0.995} &{0.99} &{0.99} &{0.995} \\
    \midrule
     \textbf{Sche. Warmup}  &{200} &{500} &{100} &{200} &{200} \\
    \midrule
    \textbf{Total Epoch}  &{50} &{50} &{50} &{50} &{50} \\
    \midrule
    \textbf{Pre-training Epoch}  &{1} &{2} &{12} &{5} &{6} \\
    \midrule
    \textbf{$\sigma_{co}$ in Eq.5 of UTL}            & {0.9} & {0.9} & {0.9} & {0.9} & {0.9} \\
    \midrule
    \textbf{$\sigma_{ua}$ in Eq.5 of UTL}            & {0.01} & {0.05} & {0.1} & {0.2} & {0.2} \\
    \midrule
    \textbf{$K$ in Eq.2 of UTL}   & {8} & {8} & {8} & {8} & {8} \\
    \midrule
    \textbf{Dropout Rate}   & {0.5} & {0.5} & {0.5} & {0.5} & {0.5} \\
    \midrule
    \textbf{\makecell[c]{ratio $\delta$ of SCL}}            & {0.3} & {0.4} & {0.3} & {0.1} & {0.1} \\
    \midrule
    \textbf{\makecell[c]{Update Cycle\\ (iterations)}}            & {6000} & {7240} & {300} & {2000} & {3200}\\
    \bottomrule
    \end{tabular}
    \caption{Hyperparameters on five DS-NER datasets.
    UTL means Uncertainty-Aware Teacher Learning and SCL means Student-Student Collaborative Learning.}
    \label{tab_hyper}
% \vspace{-0.1in}
\end{table}

\begin{table}
\centering
% \resizebox{\linewidth}{!}{%
\small
\begin{tabular}{lcccc}
\toprule
\textbf{Ratio}  & ATSEN & SCDL & BOND & \textbf{Ours} \\
\midrule
\textbf{10\% \quad \quad} & \underline{90.19} & 90.15 & 87.63 & \textbf{90.38}\\
\textbf{20\% \quad \quad} & \underline{90.03} & 89.85 & 88.03 & \textbf{90.22}\\
% \textbf{0.2} & / & / & / & / \\
\textbf{30\% \quad} & \underline{89.79} & 89.48 & 86.80 & \textbf{89.88} \\
\textbf{40\% \quad} & \underline{88.97} & 88.49 & 84.42 & \textbf{89.11} \\
% \textbf{0.4} & / & / & / & / \\
\textbf{50\% \quad} & \underline{84.77} & 83.66 & 82.56 & \textbf{86.27} \\
\textbf{60\% \quad} & {82.55} & \underline{82.64} & 80.94 & \textbf{84.96} \\
% \textbf{0.6} & / & / & / & / \\
\textbf{70\% \quad} & 75.75 & 76.88 & \underline{77.38} & \textbf{80.66} \\
\textbf{80\% \quad} & \underline{56.61} & 55.26 & 50.49 & \textbf{59.80} \\
% \textbf{0.8} & / & / & / & / \\
\textbf{90\% \quad}  & \underline{19.59} & 17.09 & 14.85 & \textbf{22.26} \\
\bottomrule
\end{tabular}
% }
\caption{F1 on CoNLL03 with different noise ratios.}
\label{tab_noise}
\end{table}

\begin{table}[h]
  \centering
  \small
% \resizebox{}{}{%
\begin{tabular}{lccc}
\toprule
\textbf{$\theta_{ua}$} & \textbf{P} & \textbf{R} & \textbf{F1} \\
\midrule
\textbf{-w/o UTL} & 86.56 & 84.37 & 85.45 \\
 \textbf{0.000 \quad \quad \quad \quad \quad } & 00.00 & 00.00 & 00.00 \\
\textbf{0.001}\quad\quad\quad & 00.00 & 00.00 & 00.00 \\
\textbf{0.005} & 85.65 & 82.68 & 84.14 \\
 \textbf{0.010} & \textbf{87.33} & \textbf{85.90} & \textbf{86.61} \\
% \rowcolor{red!5} \textbf{0.050} & - & - & - \\
% \rowcolor{red!5} \textbf{0.100} & - & - & - \\
\textbf{0.500} & 87.22 & 84.71 & 85.95 \\
\textbf{0.800} & 87.60 & 85.06 & 86.32 \\
\textbf{1.000} & 87.27 & 85.56 & 86.41 \\
\textbf{10.00} & 87.27 & 85.56 & 86.41 \\
\textbf{100.0} & 86.56 & 84.37 & 85.45 \\
\textbf{1,000} & 86.56 & 84.37 & 85.45 \\
% \textbf{10.00} & 87.27 & 85.56 & 86.41 \\
\bottomrule
\end{tabular}%
% }
\caption{F1 on CoNLL03 with different threshold $\sigma_{ua}$ in Uncertainty-Aware Label Selection. 
UTL means Uncertainty-Aware Teacher Learning.}
\label{tab_ua}
\end{table}

% \begin{table}[h]
%   \centering
%   \small
% % \resizebox{}{}{%
% \begin{tabular}{lccc}
% \toprule
% \textbf{$\theta_{ua}$} & \textbf{P} & \textbf{R} & \textbf{F1} \\
% \midrule
% \textbf{-w/o UTL} & 86.56 & 84.37 & 85.45 \\
% \rowcolor{red!5} \textbf{2} & 87.60 & 85.06 & 86.32 \\
% \rowcolor{red!5} \textbf{4} & 87.27 & 85.56 & 86.41 \\
% \rowcolor{red!5} \textbf{8} & 87.27 & 85.56 & 86.41 \\
% \rowcolor{red!5} \textbf{10} & 87.27 & 85.56 & 86.41 \\
% \rowcolor{red!5} \textbf{10} & 87.27 & 85.56 & 86.41 \\
% % \textbf{100.0} & 86.56 & 84.37 & 85.45 \\
% % \textbf{1,000} & 86.56 & 84.37 & 85.45 \\
% % \textbf{10.00} & 87.27 & 85.56 & 86.41 \\
% \bottomrule
% \end{tabular}%
% % }
% \caption{F1 on CoNLL03 with different hyperparameter $K$ in Uncertainty Estimation. 
% UTL means Uncertainty-Aware Teacher Learning.}
% \label{tab_kkk}
% \end{table}

\begin{table}[!h]
  \centering
  \small
% \resizebox{}{}{%
\begin{tabular}{lccc}
\toprule
\textbf{$K$} & \textbf{P} & \textbf{R} & \textbf{F1} \\
\midrule
\textbf{-w/o SCL} & 86.44 & 83.98 & 85.19 \\
\textbf{0.1 \quad \quad \quad} & 86.81 & 84.92 & 85.85 \\
 \textbf{0.2} & \textbf{87.35} & 84.33 & 85.82 \\
\textbf{0.3} & 87.33 & \textbf{85.90} & \textbf{86.61} \\
 \textbf{0.4} & 86.95 & 84.58 & 85.75 \\
\textbf{0.5} & 86.28 & 84.41 & 85.33 \\
\textbf{0.8} & 86.27 & 84.01 & 85.13 \\
 \textbf{1.0} & 85.70 & 83.68 & 84.68 \\
\bottomrule
\end{tabular}%
% }
\caption{F1 on CoNLL03 with different ratio $\delta$ of selected labels in Student-Student Collaborative Learning.
SCL means Student-Student Collaborative Learning.
}
\label{tab_cl}
\end{table}

\section{Parameter Study}
\label{appendix_cparameter}
In Figure \ref{fig:UA} and Table \ref{tab_ua}, we analyze the impact of $\sigma_{ua}$ in Eq.\ref{eql:mask_matrix} within Uncertainty-Aware Label Selection.
Notably, for minimal values of $\sigma_{ua}$, such as 0 and 0.001, the Uncertainty-Aware Label Selection phase filters and masks all samples. Consequently, the student network becomes incapable of parameter updates, rendering the entire teacher-student framework non-trainable.
When the parameter $\sigma_{ua}$ is in a reasonable interval, the effectiveness of the model is always improved due to the inclusion of filtered reliable labels in the self-training stage.
Ultimately, when $\sigma_{ua}$ reaches an excessive magnitude, the filtering capacity of the Uncertainty-Aware Label Selection stage is nullified, rendering the outcome akin to Uncertainty-Aware Teacher Learning omission. 
Therefore, while using different values of  $\sigma_{ua}$ tends to improve the performance, choosing $\sigma_{ua}$ wisely and rationally is crucial for optimizing Uncertainty-Aware Teacher Learning.
In Figure \ref{fig:Coteaching} and Table \ref{tab_cl}, we also explore the impact of the ratio $\delta$ of selected labels in Student-Student Collaborative Learning. 
A small $\delta$ enables the student network to partially leverage reliable labels from its counterpart, resulting in improved outcomes compared to scenarios without such collaborative learning. 
As $\delta$ increases, the transfer of these reliable labels diminishes the likelihood of learning incorrect labels from teacher-generated pseudo labels, thereby enhancing overall performance.
Conversely, an excessively large $\delta$ adversely affects performance. This is attributed to the pseudo labels of selected samples, which, with a high transfer proportion (e.g., $\delta = 0.8$), cease to qualify as small-loss samples and are more prone to containing noise. 
Hence, proportion selection of $\delta$ proves critical for optimizing the efficacy of Student-Student Collaborative Learning.

\section{Difference between Previous Methods}
We will carefully compare previous methods to explain our motivation and the differences between previous methods and our proposed components.

\paragraph{Uncertainty-Aware Teacher Learning}

% Due to the noisy samples from DS-NER dataset, which may cause the models inevitably overfit and result in poor performance, teacher-student Framework is widely used in previous works to select reliable labels and reduce the distribute of the label noise. These works generally assumed that predictions with high confidence scores tend to be correct. Specifically, these methods uses pseudo labels with high-confidence scores as "ground truth" for the optimization of student models. However, based on our observation, the overlooked incorrect high-confidence pseudo labels are widely spread in the previous methods.
% It is clear that training with these incorrect high-confidence pseudo labels could negatively impact the performance of the student models. To address this issue, we propose an Uncertainty-Aware Teacher Learning method, which includes Uncertainty Estimation and Uncertainty-Aware Label Selection components.Our approach tackles the issue from an uncertainty perspective, postulating that samples with lower uncertainty are more likely to be accurately labeled. By leveraging prediction uncertainty, the Uncertainty-Aware Teacher Learning method attempts to reduce the number of incorrect pseudo labels.

Most research on uncertainty estimation focuses on computer vision because it provides visual validation on uncertainty quality. 
For example, \citet{DBLP:conf/iclr/RizveDRS21} first introduces uncertainty to filter the low-quality labels in the semi-supervised image classification task. 
However, very little research about uncertainty has been presented in the natural language process domain. 
As far as we know, we are the first to introduce the uncertainty in the DS-NER task. 
Meanwhile, different from the instance-level image classification task, the DS-NER task is based on token-level classification, which requires the model to capture the inherent token-wise label dependency.
So different from estimating uncertainty at the instance level, we analyze the unique characteristics of the DS-NER task in the paper and design Uncertainty-Aware Teacher Learning to measure uncertainty at the token level.
% In this way, we can further improve the robustness of the NER model.
On the other hand, we are the first to find that previous teacher-student methods achieved limited performance because poor network calibration produces incorrect pseudo-labeled samples in the DS-NER task.
Thus, we attempt to use uncertainty as the indicator to reduce the effect of incorrect pseudo labels within the teacher-student framework.

% Therefore, 
% Different from using uncertainty to , 

% On the other hand, the teacher model inevitably generates some noisy labels and further leads to error propagation.
% Therefore, 

% we gather the Uncertainty of token-level classification as the signal to filter the incorrect high-confidence labels. As token-level classification demands that the model grasp the intrinsic dependency of labels linked to each token, shifts in uncertainty become more pronounced and appropriate for distinguishing these pseudo labels.

% Therefore, make it more easy to filter these wrong labels and improve quality the DS-NER dataset. 

% In the field of computer vision (CV), work has been conducted on Uncertainty Awareness (UA), but this aspect hasn't been previously explored in the natural language processing (NLP) area. We are pioneers in introducing UA to DS-NER dataset. Differing from instance-level classification, we categorize UA for token-level classification as a signal to filter incorrect high-confidence labels. The token-level classification mandates that the model captures the inherent token-wise label dependency, which makes UA more conspicuous for incorrect labels. As a result, it makes it much simpler to use the UA as the identifier to filter out these incorrect labels and enhance the quality of the DS-NER dataset.
% cite

\paragraph{Student-Student Collaborative Learning}
Collaborative Learning \citep{DBLP:conf/nips/HanYYNXHTS18, DBLP:conf/icml/Yu0YNTS19, DBLP:conf/cvpr/WeiFC020} is a popular method to handle label noise, which attempts to use two different networks to provide multi-view knowledge and let them learn from each other.
\textbf{Co-teaching} \citep{DBLP:conf/nips/HanYYNXHTS18} first attempts to completely exchange reliable samples of two different networks and then update the networks by the exchanged multi-view information.
\textbf{Co-teaching+} \citep{DBLP:conf/icml/Yu0YNTS19} further proposes to use disagreement strategy to update two networks, i.e., only using prediction disagreement data from two networks to update two networks. 
\textbf{JoCoR} \citep{DBLP:conf/cvpr/WeiFC020} aims to use a designed joint loss to reduce the diversity of two networks during training and further improve the robustness of two networks.
However, these methods are designed for tasks in the computer vision area (especially image classification), and as shown in Table \ref{tab_main}, these methods often achieve limited performance in the DS-NER task.
\textbf{SCDL} designs the teacher-student framework and adopts collaborative learning in the DS-NER task.
Similar to Co-teaching, all of the pseudo labels predicted by the teacher are applied to update the noisy labels of the peer teacher-student network periodically since two teacher-student networks have different learning abilities based on different network structures. 
Different from SCDL, we aim to utilize two different student networks and let them learn from each other to reduce the negative effect of incorrect pseudo labels.
Specifically, instead of completely exchanging pseudo labels between two teachers, we allow students to transfer reliable pseudo labels and at the same time allow students to learn on their own pseudo labels generated by their teacher network.
In this way, we not only ensure that the transferred pseudo labels contain multi-view information but also ensure that the pseudo labels we transfer are high-quality by selective transfer.
Meanwhile, as the student network is updated earlier and more frequently than the teacher network, the student network is better able to capture the changes of pseudo labels than the teacher network.

\paragraph{Relation between Two Components}
% The design of Uncertainty-Aware Teacher Learning and Student-Student Collaborative Learning are not 
Designs on Uncertainty-Aware Teacher Learning and Student-Student Collaborative Learning are not independent. 
The two components can collaborate and achieve better results.
Specifically, 
(1) Uncertainty-Aware Teacher Learning can help the teacher network to generate more reliable pseudo labels and further reduce the risk of the student network updating parameters on the incorrect pseudo label.
At the same time, a more efficient student network can be achieved by learning to pseudo-label with fewer errors, which will further improve the efficiency of the Student-Student Collaborative Learning component;
(2) Based on Uncertainty-Aware Teacher Learning, the teacher network can utilize the correctly pseudo-labeled samples to alleviate the negative effect of label noise. 
However, simply masking unreliable pseudo-labeled samples can lead to underutilization of the training set, as there is no chance for the incorrect pseudo-labeled samples to be corrected and further learned.
Student-Student Collaborative Learning can allow the student network to learn from transferred reliable labels from the other student network.
Therefore, this component further enables a full exploration of mislabeled samples rather than simply filtering unreliable pseudo-labeled samples. 
Through the collaboration of the two components, as shown in Table \ref{tab_main}, CENSOR achieves the best performance among 12 baselines.
\end{document}